\documentclass{article}
\usepackage[T1]{fontenc}
\usepackage{graphicx}
\usepackage{booktabs}
\usepackage{graphicx}
\usepackage{subcaption}
\usepackage{amsmath}
\usepackage{amssymb}
\usepackage{mathtools}
\usepackage{booktabs}
\usepackage{array}
\usepackage{longtable}
\usepackage{url}

\newcolumntype{L}[1]{>{\raggedright\arraybackslash}p{#1}}
\usepackage[final]{corl_2026} 

\title{Future-Aware Flow Planning for Safe UAV Target Following}

\author{
  Boning Feng\textsuperscript{1}\quad
  Haoran Zhang\textsuperscript{2,3}\quad
  Xiaowen Bi\textsuperscript{4,5}\\
  \textbf{Yanzhen Zhang\textsuperscript{2,3}\quad Xiaodan Shi\textsuperscript{1}}\\
  \textsuperscript{1}Department of Computer and Systems Sciences,\\
  Stockholm University, Stockholm, Sweden\\
  \textsuperscript{2}School of Urban Planning and Design, Peking University,\\
  Shenzhen, 518055, China\\
  \textsuperscript{3}Guangdong Provincial Key Laboratory of Risk Perception and\\
  Sustainable Governance in Energy Transition, Shenzhen, 518055, Guangdong, China\\
  \textsuperscript{4}Department of Statistics and Data Science,\\
  Beijing Normal-Hong Kong Baptist University, Zhuhai, 519087, China\\
  \textsuperscript{5}Guangdong Provincial/Zhuhai Key Laboratory of IRADS,\\
  Beijing Normal-Hong Kong Baptist University, Zhuhai, 519087, China\\
  code: \url{https://github.com/BoningFeng/Future-Aware-Flow-Planning}
}

\begin{document}
\maketitle


\begin{abstract}
UAV target following in cluttered environments is inherently predictive: current-state followers can lag behind turns, choose blocked corridors, or trade tracking for unsafe near-horizon motion. We propose a future-aware flow planning framework for state-informed UAV target following. Predicted target futures guide clean UAV trajectory generation as horizon-aligned residual signals, while risk-scored executable-prefix repair is embedded inside the sampling loop. On fixed ID/OOD receding-horizon benchmarks, the planner improves the intended safety--tracking trade-off rather than dominating every metric: it matches zero measured ID collision rate with the highest ID safe-tracking time, and gives the lowest OOD macro collision rate and final tracking error among the displayed methods, while Future-MPC remains smoother and stronger on some thresholded OOD success metrics under its hand-designed objective. Ablations show that future adaptation improves candidate generation before safety repair, and simulator-facing stress tests probe interface, sensing, and controller-execution effects. These results support horizon-aligned future adaptation and embedded prefix repair as complementary ingredients for safe UAV target following under the tested simulation conditions.
\end{abstract}

\keywords{UAV Path Planning, Future-Aware Flow Planner, Safe Trajectory Generation} 


\section{Introduction}

\begin{figure}[t]
    \centering
    \includegraphics[width=0.76\linewidth, trim=0 55pt 0 45pt, clip]{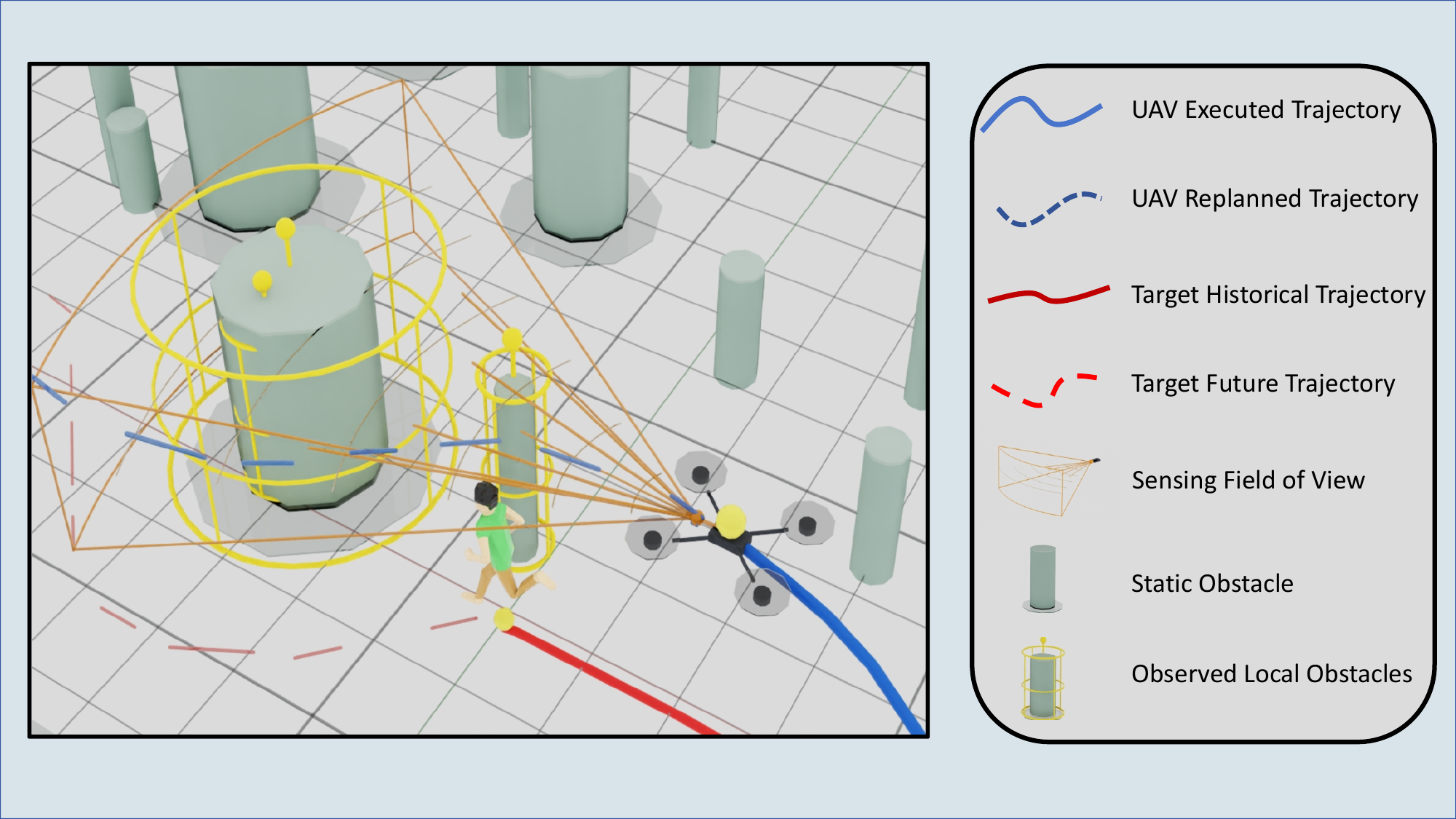}
    \caption{State-informed UAV target following: the planner conditions on recent target states, the UAV state, and ego-frame local obstacles, then executes a short safe-following prefix before replanning. The sensing field denotes the local region used for ego-frame obstacle cropping, rather than a continuous target-visibility requirement.}
    \label{fig:task_setting}
\end{figure}

UAV target following in cluttered environments requires repeated short-horizon planning under moving-target uncertainty, local obstacle observations, collision avoidance, and tight online computation constraints~\citep{chen2016tracking,han2021fasttracker,ji2022elastic,li2018uavrl,bhagat2020uav}. 
We focus on the trajectory-planning component of target following rather than full visual detection or re-identification. 
As illustrated in Fig.~\ref{fig:task_setting}, the planner receives discrete target-state observations, the current UAV state, and ego-frame local obstacles, then executes only a short prefix before replanning~\citep{zhou2021raptor,tordesillas2022faster,liu2016high,mohta2018fast,zhou2019robust}.

A central difficulty is that target following is predictive: planning only toward the current target can lag behind turns or become unsafe when clutter restricts feasible routes~\citep{chen2016tracking,wang2021visibility,gao2024probabilistic}. 
Without future-aware coupling, a learned follower can reduce instantaneous tracking error yet commit its executable prefix to a soon-obsolete corridor or react late after a target turn. 
Directly treating the predicted target path as a UAV reference is also unsafe, because the target may pass through narrow or blocked regions that a UAV should avoid. 
MPC-style systems address anticipation through receding-horizon optimization over predicted target motion~\citep{chen2016tracking,han2021fasttracker,ji2022elastic,tallamraju2018decentralized,sun2022nmpc,lee2024bpmptracker}, but they often depend on online optimization and hand-designed objectives. 
Learning-based methods can amortize behavior but are commonly reactive or task-specific~\citep{li2018uavrl,bhagat2020uav}. 
The challenge is therefore to convert uncertain target futures into UAV-feasible trajectories that preserve both tracking and executable safety.

We propose \emph{Future-Aware Flow Planning}, a learned receding-horizon trajectory generator that separates target anticipation from UAV route generation. Predicted target futures enter as horizon-aligned residual guidance for clean-trajectory flow generation, while a Risk-Scored Executable-Prefix Safety Sampler (RSEPSS) is embedded inside each sampling update so risky near-horizon segments shape generation. Fixed-case ID/OOD benchmarks and simulator-facing stress tests show that the final planner improves the safety-constrained following behavior targeted by the method: it reaches zero measured collision rate and the highest STT@8 in ID, and the lowest collision rate and best FDE on the OOD macro average, while ablations show that future adaptation improves candidate generation before RSEPSS supplies the near-horizon safety gain.


\section{Related Work}

\paragraph{Classical Replanning and Learning-Based UAV Target Following.} UAV target following has been studied through optimization-based aerial tracking, receding-horizon replanning, and learning-based control~\citep{chen2016tracking,han2021fasttracker,ji2022elastic,lee2024bpmptracker}. 
Classical tracking and MPC-style planners show the value of predicted target motion~\citep{han2021fasttracker,ji2022elastic,tallamraju2018decentralized,sun2022nmpc}, and recent aerial trackers similarly separate target-motion prediction from chasing trajectory planning~\citep{lee2024bpmptracker}. 
Quadrotor replanning methods emphasize perception-aware local planning, feasibility, and dynamics~\citep{zhou2021raptor,tordesillas2022faster,wang2022gcopter}. 
Search-, sampling-, and learning-based methods provide complementary tools, but often require online hand-designed costs or reactive task-specific policies~\citep{karaman2011sampling,kuffner2000rrtconnect,gammell2015bit,li2018uavrl,bhagat2020uav}. 
Our goal is to carry the predictive principle into a learned generative trajectory planner.

\paragraph{Generative Trajectory Planning and Clean-Trajectory Flow.} Generative planners can represent multimodal trajectories and action distributions~\citep{janner2022planning,chi2023diffusion,ze2024dp3,reuss2023beso,shaoul2025mmd}. 
Diffusion planners are expressive but costly for online replanning, whereas flow matching offers continuous transport dynamics for forecasting, generation, and robot planning~\citep{lipman2023flowmatching,ye2024tcfm,nguyen2025flowmp}. 
Because UAV target following outputs the executable trajectory itself, we use a clean-trajectory x-pred formulation~\citep{karras2022edm,li2025backtobasics,song2025l1flow}. 
Target-, intention-, and query-conditioned prediction works show the value of structured future signals~\citep{zhao2021tnt,shi2022mtr,Zhou_2023_CVPR}, but unreliable references can also mislead tracking~\citep{li2020keyfilter}; our gated residual future tokens address this tension.

\paragraph{Safety-Aware Generative Planning and Prefix Refinement.} Safety remains difficult when generative planners are deployed in closed-loop cluttered systems. 
Control barrier functions and related correction mechanisms have been incorporated into diffusion- and flow-based planners~\citep{ames2017cbf,xiao2025safediffuser,mizuta2024cobl,dai2025safeflow,yang2025safeflowmatcher,yang2025uniconflow}. 
Most operate on the full trajectory or full sampling dynamics. 
In receding-horizon UAV following, only a short prefix is executed before replanning, so RSEPSS focuses correction on risky prefix points and embeds that correction into the sampling update.

\section{Methodology}

\subsection{Problem Setup and Framework Overview}
\label{sec::3.1}

At each replanning step, the UAV observes local obstacles, its motion state, and recent target history, then generates a target-following trajectory and executes only its first prefix before replanning~\citep{liu2016high,mohta2018fast,zhou2021raptor,wu2022perceptionaware,tordesillas2022panther}.

\begin{figure}[t]
    \centering
    \includegraphics[width=0.96\textwidth]{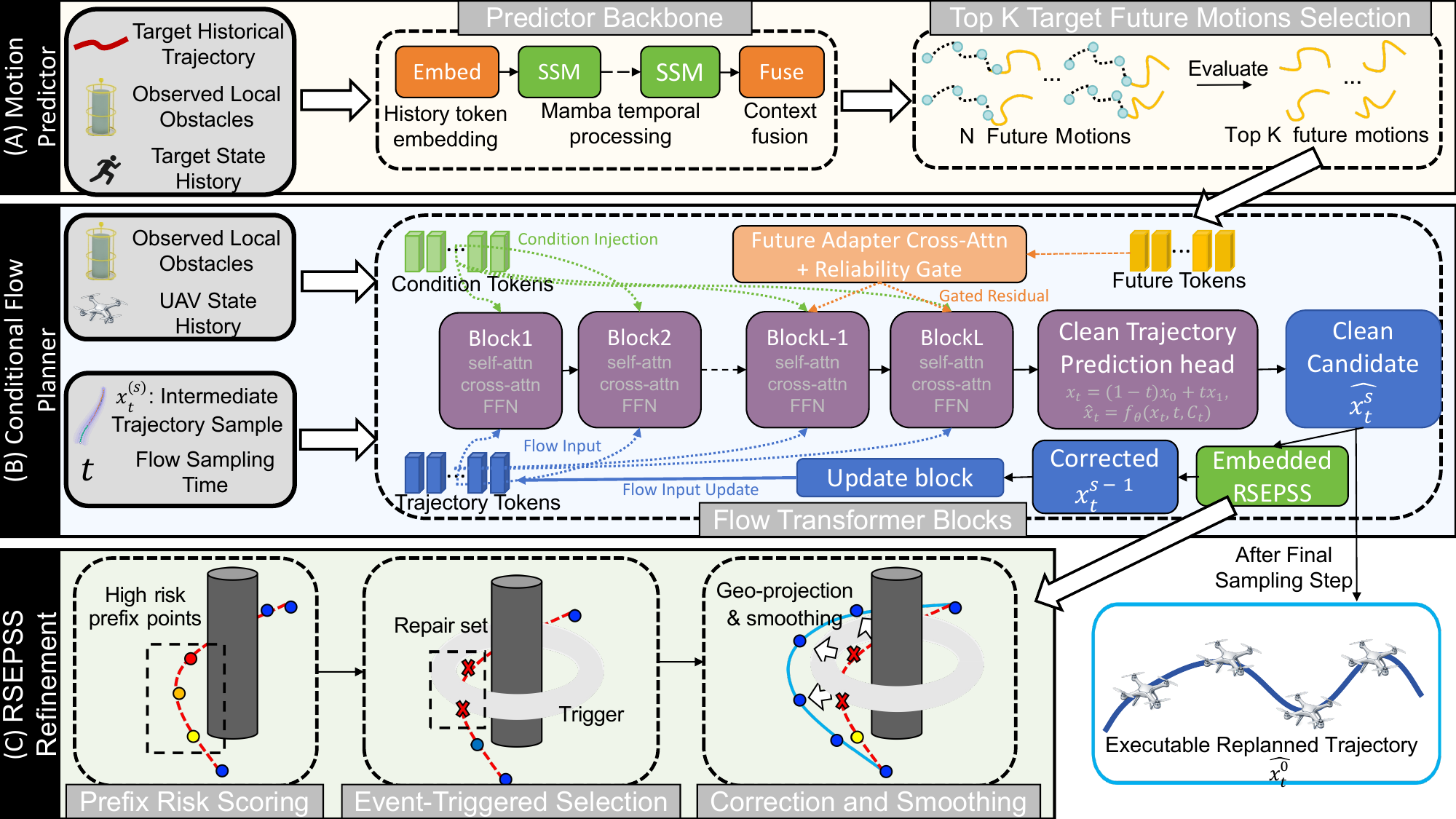}
    \caption{Overview of the proposed framework, which couples target-future prediction, gated future-adapter flow planning, and embedded executable-prefix safety refinement. The predicted target future enters the planner as gated residual guidance in selected flow blocks, while RSEPSS repairs executable prefixes inside the sampling loop rather than only after final trajectory generation.}
    \label{fig:method_framework}
\end{figure}

Fig.~\ref{fig:method_framework} summarizes the framework: a target predictor estimates short-horizon futures, a gated future-adapter flow planner uses them as horizon-aligned signals, and RSEPSS corrects risky prefix points during sampling. Future target motion influences the horizon plan, but the generated object remains a UAV-feasible trajectory rather than a copied target path. Let $s_t^{\mathrm{uav}}$ be the current UAV state, $\tilde{o}_t=\{o_i\}_{i=1}^{M}$ the ego-frame obstacle cloud, $h_t^{\mathrm{tar}}$ the target history, and $\hat{Y}_t^{\mathrm{tar}}=\{\hat{y}_{t+1}^{\mathrm{tar}},\ldots,\hat{y}_{t+H}^{\mathrm{tar}}\}$ the predicted target future. The planner condition is
\begin{equation}
c_t = \bigl(s_t^{\mathrm{uav}}, \tilde{o}_t, h_t^{\mathrm{tar}}, \hat{Y}_t^{\mathrm{tar}}\bigr).
\end{equation}
The planner outputs $\tau_t=\{p_{t+1},\ldots,p_{t+H}\}$. In the conditional flow sampler, $x_{t_s}\in\mathbb{R}^{H\times 3}$ is an intermediate trajectory sample. Each step predicts a clean candidate $\hat{x}_1^{(s)}$, applies RSEPSS to obtain $\tilde{x}_1^{(s)}$, and updates the flow state. The final corrected prefix is executed before the next replan.

\subsection{Target Motion Predictor}
\label{sec:target_predictor}

Target following requires estimating how the target may move over the next horizon, because the executed prefix should already account for upcoming target turns. 
We use a lightweight predictor to produce short-horizon target-motion hypotheses; these guide tracking but do not determine the UAV route.

Given the recent target history $h_t^{\mathrm{tar}}$, the local obstacle point cloud $\tilde{o}_t$, the UAV state $s_t^{\mathrm{uav}}$, the current target state, and a future time grid, the predictor first embeds the target-history tokens and fuses them with the scene and motion context. 
A Mamba-based temporal backbone then processes the history sequence and outputs $K$ future target hypotheses with confidence logits~\citep{gu2023mamba},
\begin{equation}
\left\{
\hat{Y}_{t}^{(k)}, \ell_t^{(k)}
\right\}_{k=1}^{K}
=
P_\phi\!\left(h_t^{\mathrm{tar}},\tilde{o}_t,s_t^{\mathrm{uav}},y_t^{\mathrm{tar}},v_t^{\mathrm{tar}}\right),
\qquad
\hat{Y}_{t}^{(k)}=\{\hat{y}_{t+1}^{(k)},\ldots,\hat{y}_{t+H}^{(k)}\}.
\end{equation}
Here $\ell_t^{(k)}$ scores the $k$-th target-future hypothesis and each $\hat{Y}_{t}^{(k)}$ is represented in the current UAV ego frame. 
We append the relative future timestamp to each predicted target position to form temporal future features $F_t^{(k)}=\{[\hat{y}_{t+i}^{(k)},\Delta t_i]\}_{i=1}^{H}$.

At inference time, the top-ranked hypotheses and logits are passed to the planner, preserving temporal correspondence while allowing obstacle-aware detours.

\subsection{Gated Future-Adapter Conditional Flow Planner}
\label{sec:flow_planner}

The predicted target path should not be treated as a UAV trajectory because obstacles, dynamics, or prediction errors may require a different route. 
We therefore formulate planning as conditional clean-trajectory flow generation: the planner predicts UAV trajectories directly, while target futures resolve horizon-level tracking ambiguity before the prefix is executed.

At sampling step $s$, the planner receives an intermediate full-horizon trajectory sample $x_{t_s}\in\mathbb{R}^{H\times 3}$, the flow time $t_s$, and the planning context $c_t$. 
The trajectory sample is embedded as
\begin{equation}
H_0 = E_x(x_{t_s}) + E_t(t_s) + E_{\mathrm{pos}},
\end{equation}
where $E_x$ is a pointwise trajectory embedding, $E_t$ is a flow-time embedding, and $E_{\mathrm{pos}}$ is the horizon-position embedding. 
In parallel, a scene encoder maps the local obstacle point cloud $\tilde{o}_t$ into obstacle tokens, state-history encoders map the UAV state and recent motion information into motion-context tokens, and a future encoder maps the selected target-motion hypotheses into temporal future tokens. 
These tokens provide obstacle geometry, local motion context, and target-motion anticipation to the planner while keeping the generated object as a UAV trajectory.

The main backbone consists of $L$ Flow Transformer blocks. 
Each block updates the trajectory tokens through trajectory self-attention, cross-attention to condition tokens, and feed-forward layers,
\begin{equation}
H_l = B_l\!\left(H_{l-1}; Z_t^{\mathrm{obs}}, Z_t^{\mathrm{mot}}, Z_t^{\mathrm{fut}}\right),
\qquad l=1,\ldots,L,
\end{equation}
where $Z_t^{\mathrm{obs}}$, $Z_t^{\mathrm{mot}}$, and $Z_t^{\mathrm{fut}}$ denote obstacle, motion-context, and target-future tokens, respectively.

We train the planner with a clean-trajectory prediction objective. 
Given a clean expert UAV trajectory $x_1$ and a random initial trajectory $x_0$, we construct an interpolated flow state
\begin{equation}
x_\lambda = (1-\lambda)x_0 + \lambda x_1,
\qquad
\lambda\sim\mathcal{U}(0,1).
\end{equation}
Rather than predicting a velocity or noise target, the planner predicts the clean trajectory endpoint,
\begin{equation}
\mathcal{L}_{\mathrm{xpred}}
=
\mathbb{E}_{x_0,x_1,\lambda,c_t}
\left[
\left\|
f_\theta(x_\lambda,\lambda,c_t)-x_1
\right\|_2^2
\right].
\end{equation}
At inference time, the sampler evaluates the same network at discrete flow times $t_s$. 
The final hidden tokens are decoded by an x-prediction head to produce a clean candidate trajectory,
\begin{equation}
\hat{x}_1^{(s)}
=
f_\theta(x_{t_s},t_s,c_t)
=
x_{t_s}+D_\theta(H_L),
\end{equation}
where $D_\theta$ denotes the clean-trajectory prediction head.

To refine how target futures affect generation, we add a gated future-adapter to late Flow Transformer blocks. 
The adapter encodes the selected target future into horizon-aligned tokens $Z_t^{\mathrm{aux}}$ and injects them only as a gated residual refinement, avoiding hard coordinate-level anchoring. 
For a late block $l$, it computes a residual feature by cross-attending from planner hidden tokens to these auxiliary future tokens,
\begin{equation}
R_l =
A_l\!\left(
\mathrm{CrossAttn}
\left(
\mathrm{LN}(H_l),
\mathrm{LN}(Z_t^{\mathrm{aux}}),
\mathrm{LN}(Z_t^{\mathrm{aux}})
\right)
\right),
\end{equation}
where $A_l$ is a lightweight bottleneck adapter. 
A reliability gate controls the residual strength,
\begin{equation}
H_l' = H_l + \alpha_l g_l R_l,
\qquad
g_l =
\sigma\!\left(
G_l(\mathrm{Pool}(Z_t^{\mathrm{aux}}))
\right),
\end{equation}
where $\alpha_l$ is a small learnable scale. 
The gate is learned from the future-token summary, so the adapter can modulate the strength of predictor-derived residual features rather than passing them to the planner with a fixed weight.

The adapter is deliberately residual and late-stage: it lets target predictions change tracking intent while preserving the obstacle-aware route structure produced by the base planner. 
The resulting clean candidate $\hat{x}_1^{(s)}$ is passed to RSEPSS.

\subsection{Embedded RSEPSS Sampling}
\label{sec:embedded_rsepss}

The clean candidate $\hat{x}_1^{(s)}$ may still contain risky near-horizon points. 
Post-sampling repair is one-shot and cannot affect later sampling states, which can leave a mismatch between the corrected prefix and the remaining trajectory. 
We therefore embed executable-prefix correction into the sampling loop so each corrected clean candidate participates in the next flow update.

At sampling step $s$, let
\begin{equation}
\hat{x}_1^{(s)}=\{\hat{q}_1,\hat{q}_2,\ldots,\hat{q}_H\}
\end{equation}
denote the clean candidate trajectory predicted by the planner, and let $\tilde{o}_t=\{o_i\}_{i=1}^{M}$ denote the local obstacle point cloud. 
RSEPSS extracts the executable prefix
\begin{equation}
\hat{x}_{1,\mathrm{pre}}^{(s)}=\{\hat{q}_1,\ldots,\hat{q}_{H_{\mathrm{exec}}}\},
\end{equation}
and evaluates the nearest-obstacle distance for each prefix point,
\begin{equation}
d_j = \min_i \|\hat{q}_j-o_i\|_2,\qquad j=1,\ldots,H_{\mathrm{exec}} .
\end{equation}
A prefix point is considered risky when its clearance falls below a trigger margin. 
We define a temporally weighted risk score
\begin{equation}
r_j = w_j [d_{\mathrm{trig}}-d_j]_+ ,
\end{equation}
where $[\cdot]_+$ is the positive part and $w_j$ assigns larger weight to earlier executable points.

When the maximum prefix risk exceeds a threshold, RSEPSS selects a sparse repair set from high-risk points and their temporal neighbors. 
For each selected point, it computes an outward correction direction from the nearest obstacle point,
\begin{equation}
u_j =
\frac{\hat{q}_j-o_j^{\mathrm{nn}}}
{\|\hat{q}_j-o_j^{\mathrm{nn}}\|_2+\epsilon},
\end{equation}
where $o_j^{\mathrm{nn}}$ is the nearest obstacle point. 
The point is then projected away from the obstacle until the desired local safety margin is approached. 
After geometric correction, a lightweight smoothing step is applied on the repaired prefix to reduce abrupt local changes while keeping the start point consistent with the current UAV state.

The corrected prefix is written back into the clean candidate trajectory, producing
\begin{equation}
\tilde{x}_1^{(s)}
=
\mathrm{RSEPSS}\!\left(\hat{x}_1^{(s)},\tilde{o}_t\right).
\end{equation}
The flow sampler then uses this corrected clean candidate, rather than the raw prediction, to advance the sampling state:
\begin{equation}
x_{t_{s-1}}
=
\mathrm{Update}\!\left(x_{t_s},\tilde{x}_1^{(s)},t_s,t_{s-1}\right).
\end{equation}
Thus, safety correction affects the next flow state at every sampling step. 
After the final sampling step, the corrected executable prefix is sent to the UAV for execution, and the same procedure is repeated at the next replanning time.

Thus, RSEPSS is part of trajectory generation rather than a detached output filter. 
It modifies only the executable prefix, matching the receding-horizon interface while leaving the longer horizon available for target-aware replanning.
\section{Experiments}

\begin{figure*}[t]
    \centering
    \begin{subfigure}[t]{0.24\linewidth}
        \caption{SafeFlow}
        \includegraphics[width=\linewidth]{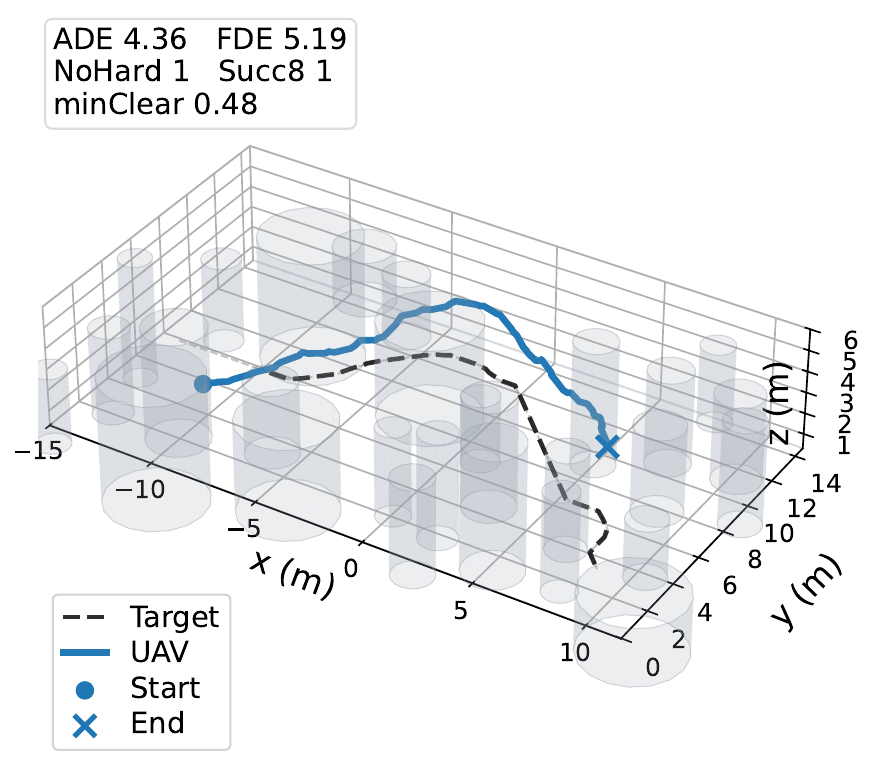}
        \label{fig:id-low-safeflow}
    \end{subfigure}
    \hfill
    \begin{subfigure}[t]{0.24\linewidth}
        \caption{SafeFlowMatcher}
        \includegraphics[width=\linewidth]{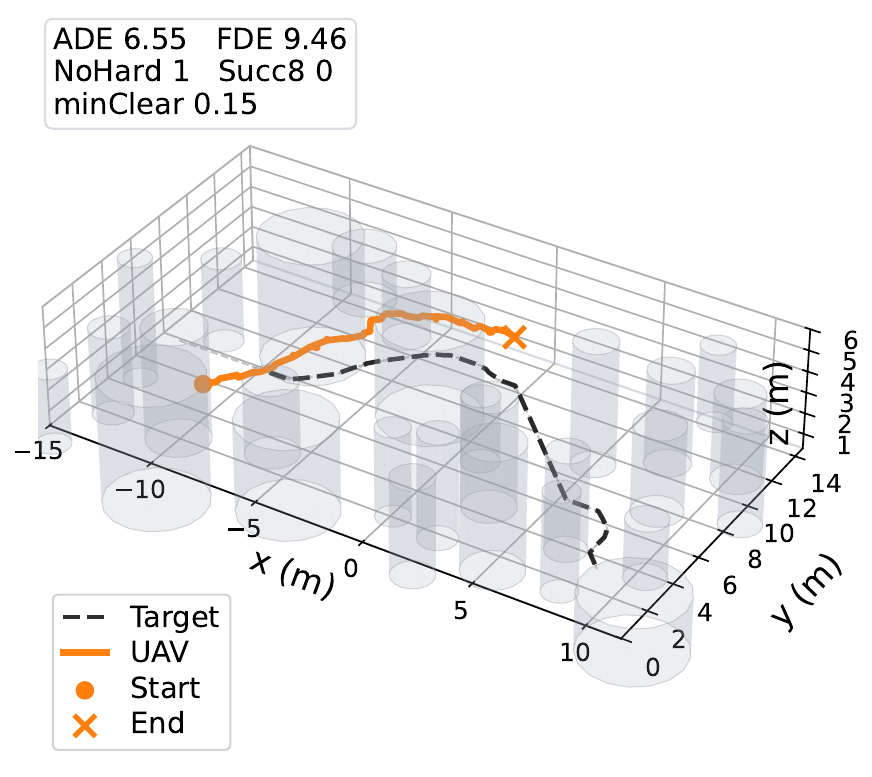}
        \label{fig:id-low-safeflowmatcher}
    \end{subfigure}
    \hfill
    \begin{subfigure}[t]{0.24\linewidth}
        \caption{Future-MPC}
        \includegraphics[width=\linewidth]{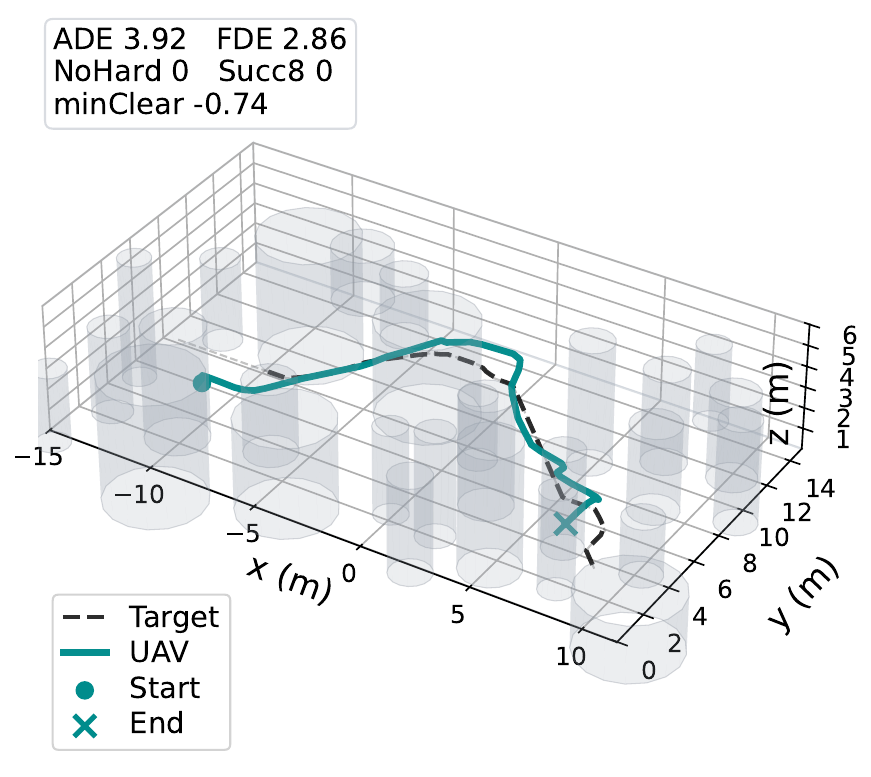}
        \label{fig:id-low-future-mpc}
    \end{subfigure}
    \hfill
    \begin{subfigure}[t]{0.24\linewidth}
        \caption{Ours}
        \includegraphics[width=\linewidth]{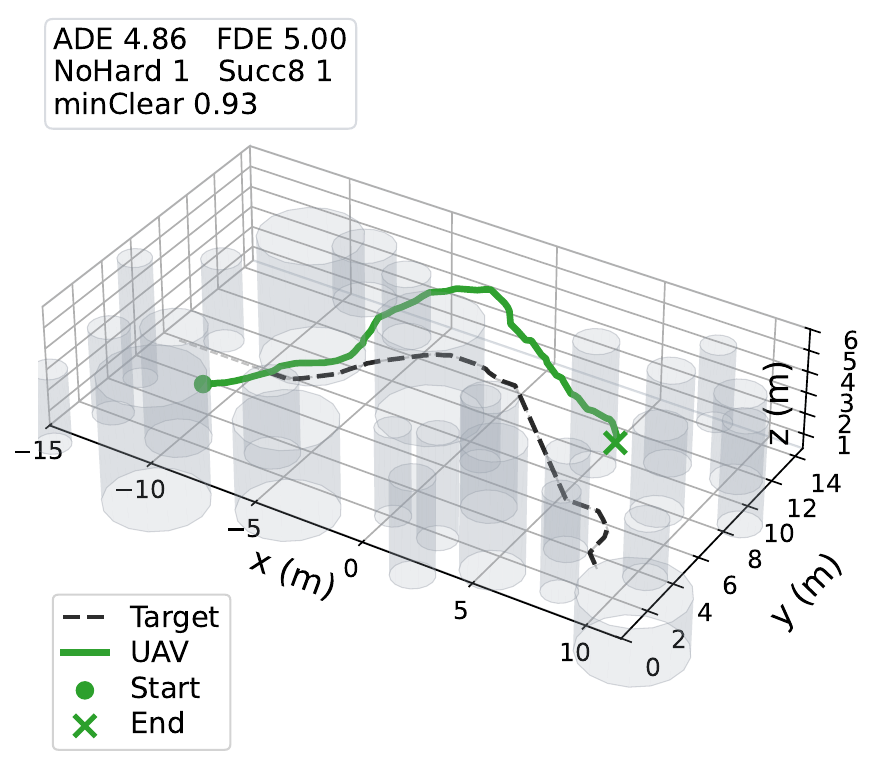}
        \label{fig:id-low-ours}
    \end{subfigure}
    \caption{Qualitative comparison on an ID low-density scene. The aligned view highlights how the compared planners trade raw target tracking, clearance, and safe executable prefixes.}
    \label{fig:id-low-main-qual}
\end{figure*}

\subsection{Evaluation Questions}
\label{sec:setup}

We evaluate the planner along three axes. The ID/OOD benchmark tests whether future-aware generation and embedded prefix repair improve the closed-loop tracking--safety trade-off relative to safety-aware learned flow baselines and a fixed Future-MPC reference. The fixed set contains 150 effective ID forest episodes and 200 OOD episodes, with all methods sharing the horizon, executed prefix, obstacle budget, target predictor, and metrics. Adapter and predictor analyses test whether target futures change the candidate UAV trajectory before safety repair. Finally, a flight-controller-in-the-loop validation streams the planned reference through Isaac Sim, Pegasus, and PX4 SITL~\citep{nvidia2026isaacsim,pegasus2024simulator,meier2015px4}. Full benchmark definitions, controlled variables, metrics, statistical uncertainty, and simulator protocols are in Appendix~\ref{app:setup}; implementation and training details are in Appendix~\ref{app:implementation_training}.

\begin{table}[t]
\centering
\caption{ID macro summary. Values are averaged over low-, medium-, and high-density forest groups. Full per-density results are in Appendix~\ref{app:detailed_main_tables}, and the uncertainty protocol is in Appendix~\ref{app:statistical_stability}. Smoothness is jerk RMS.}
\label{tab:id_main}
\scriptsize
\setlength{\tabcolsep}{2.4pt}
\renewcommand{\arraystretch}{1.05}
\resizebox{\linewidth}{!}{
\begin{tabular}{lccccccc}
\toprule
\textbf{Method} & \textbf{ADE}$\downarrow$ & \textbf{FDE}$\downarrow$ & \textbf{CR}$\downarrow$ & \textbf{Smooth.}$\downarrow$ & \textbf{STT@8}$\uparrow$ & \textbf{STR@8}$\uparrow$ & \textbf{Time}$\downarrow$ \\
\midrule
SafeFlow        & 4.128 & 5.622 & 0.006 & 103.116 & 0.938 & \textbf{0.933} & 0.213 \\
SafeFlowMatcher & 6.042 & 8.747 & \textbf{0.000} & 113.782 & 0.740 & 0.469 & \textbf{0.046} \\
Future-MPC      & \textbf{3.881} & \textbf{3.534} & 0.043 & \textbf{27.555} & 0.583 & 0.583 & 0.125 \\
Ours            & 4.365 & 5.448 & \textbf{0.000} & 82.094 & \textbf{0.960} & 0.913 & 0.300 \\
\bottomrule
\end{tabular}}
\end{table}

\begin{table}[t]
\centering
\caption{OOD macro summary. Values are averaged over shape, distribution, speed, and combined shifts. Full per-shift results are in Appendix~\ref{app:detailed_main_tables}, and the uncertainty protocol is in Appendix~\ref{app:statistical_stability}. Smoothness is jerk RMS.}
\label{tab:ood_main}
\scriptsize
\setlength{\tabcolsep}{2.4pt}
\renewcommand{\arraystretch}{1.05}
\resizebox{\linewidth}{!}{
\begin{tabular}{lccccccc}
\toprule
\textbf{Method} & \textbf{ADE}$\downarrow$ & \textbf{FDE}$\downarrow$ & \textbf{CR}$\downarrow$ & \textbf{Smooth.}$\downarrow$ & \textbf{STT@8}$\uparrow$ & \textbf{STR@8}$\uparrow$ & \textbf{Time}$\downarrow$ \\
\midrule
SafeFlow        & \textbf{3.518} & 3.926 & 0.014 & 110.019 & 0.814 & 0.812 & 0.199 \\
SafeFlowMatcher & 5.526 & 7.177 & 0.010 & 121.767 & 0.733 & 0.562 & \textbf{0.043} \\
Future-MPC      & 4.770 & 4.288 & 0.003 & \textbf{27.226} & \textbf{0.958} & \textbf{0.958} & 0.124 \\
Ours            & 3.653 & \textbf{3.719} & \textbf{0.002} & 89.830 & 0.896 & 0.854 & 0.287 \\
\bottomrule
\end{tabular}}
\end{table}

\subsection{Closed-Loop Tracking--Safety Trade-off}
\label{sec:main_results}

Table~\ref{tab:id_main} tests the nominal forest distribution, with full per-density results in Appendix~\ref{app:detailed_main_tables}. The ID evidence supports the main claim: the proposed method is the safest displayed planner under the nominal benchmark. It matches the best measured collision rate (0.000) and reaches the highest STT@8 (0.960), improving the safe-tracking objective targeted by the method. Future-MPC illustrates the different profile of a fixed receding-horizon objective: it gives lower ADE/FDE and jerk, but its collision rate and safe-tracking time degrade in dense clutter (CR 0.043, STT@8 0.583). Fig.~\ref{fig:id-low-main-qual} makes the trade-off visible: the planner accepts some endpoint-error cost to preserve an executable safe prefix near obstacles.

Table~\ref{tab:ood_main} then probes whether the same behavior survives shifted obstacle geometry, obstacle distribution, target speed, and their combination. The OOD results show that no method dominates all metrics, but the proposed method retains the advantages most aligned with safe learned target following: it gives the lowest collision rate (0.002) and best average FDE (3.719) among the displayed methods. The fixed Future-MPC reference favors lower jerk and thresholded STT@8/STR@8 under its explicit online objective, but it does not dominate tracking accuracy or safety. This bounds the claim: the learned planner improves safety-constrained generation relative to learned flow baselines and remains competitive with a hand-designed receding-horizon reference, especially on collision avoidance and final target tracking under shift. OOD visualizations and per-shift breakdowns in Appendix~\ref{app:ood_qualitative} and Appendix~\ref{app:detailed_main_tables} show where the regimes separate, while Appendix~\ref{app:robustness} gives a conditional risk-decomposition view of these OOD trends without claiming a universal safety certificate.

\subsection{Flight-Controller-in-the-Loop Validation}
\label{sec:simulator_validation}

\begin{figure}[t]
\centering
\includegraphics[width=0.65\linewidth]{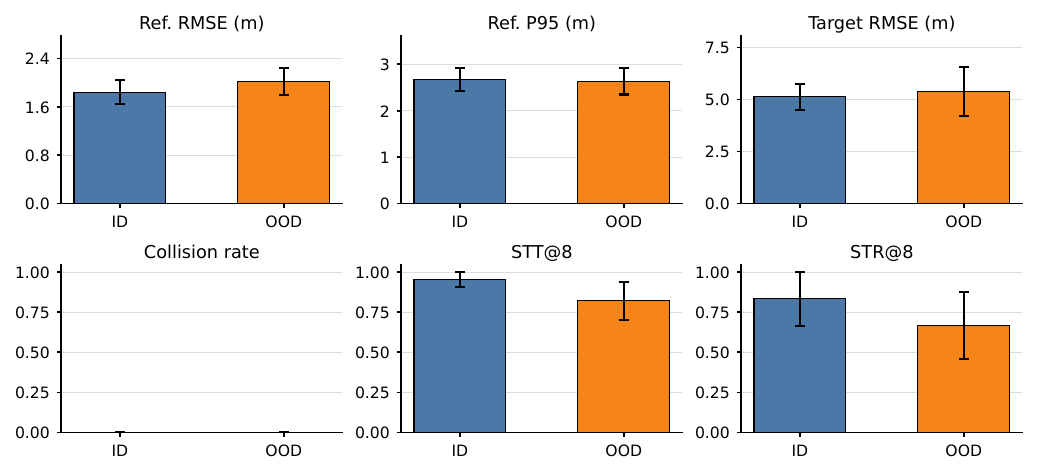}
\caption{PX4 SITL flight-controller-in-the-loop validation over 120 runs. Bars show ID forest-medium and OOD combined-shift wall means; error bars are standard errors over scenes. }
\label{fig:px4_fcitl_bars}
\end{figure}

The main rollouts isolate planner behavior under a fixed planning interface. To test whether the generated references remain usable by a realistic flight-control stack, we run 120 PX4 SITL offboard executions through Isaac Sim and Pegasus: sixty ID forest-medium scenes and sixty OOD combined-shift wall scenes. The planner reference is streamed as local setpoints to PX4 at 20 Hz; each run uses 36 reference points after the simulated Iris vehicle arms and enters offboard mode. As summarized in Fig.~\ref{fig:px4_fcitl_bars}, PX4 reference tracking remains stable in both groups (RMSE $1.84\pm0.20$ m in ID and $2.02\pm0.22$ m in OOD, mean $\pm$ SE), with zero hard-collision rate in all 120 runs. Path-aligned task tracking remains high but not saturated (STT@8/STR@8: 0.95/0.83 in ID and 0.82/0.67 in OOD), which keeps the claim at flight-control compatibility rather than physical-flight validation. PX4 bootstrap intervals are reported in Appendix~\ref{app:statistical_stability}; lighter L1/L2 simulator-facing stress tests and a representative Isaac Sim rollout visualization (Fig.~\ref{fig:app_sim_exp}) are reported in Appendix~\ref{app:sim_stress}.

\subsection{Effect of the Gated Future Adapter}
\label{sec:gated_future_adapter_effect}

\subsubsection{Main effect on candidate generation}

Future conditioning is useful only if the predictor is accurate enough and the planner can use it without copying the target path. A held-out probe shows future-target ADE/FDE below 0.28/0.65 m on simple and complex motion bins (Appendix~\ref{app:predictor_quality}). Table~\ref{tab:gfa_main_effect} removes RSEPSS and compares the base pooled future-conditioned planner with the residual adapter under the same predictor, interface, training budget, and rollout settings. This isolates the main future-aware question: weakly pooled future information can still leave the candidate generator prone to late or locally risky tracking, while the adapter aligns the UAV route with upcoming target motion. The adapter reduces ID ADE/FDE from 4.628/5.032 to 3.462/4.453 and OOD ADE/FDE from 5.243/4.822 to 3.002/2.698; it also lowers collision rate and improves safe-tracking metrics before explicit safety repair. Same-scene supplementary videos compare the full model with the no-future/no-adapter control, illustrating the closed-loop effect of future-aware coupling after target turns.

\begin{table}[t]
\centering
\caption{Main effect of the gated future adapter without RSEPSS. The control is the base pooled future-conditioned planner without the auxiliary residual future-adapter pathway.}
\label{tab:gfa_main_effect}
\scriptsize
\setlength{\tabcolsep}{3.0pt}
\renewcommand{\arraystretch}{1.05}
\resizebox{\linewidth}{!}{
\begin{tabular}{llccccccc}
\toprule
\textbf{Split} & \textbf{Variant} & \textbf{ADE}$\downarrow$ & \textbf{FDE}$\downarrow$ & \textbf{CR}$\downarrow$ & \textbf{Smooth.}$\downarrow$ & \textbf{STT@8}$\uparrow$ & \textbf{STR@8}$\uparrow$ & \textbf{Time}$\downarrow$ \\
\midrule
ID
& base pooled future-conditioned & 4.628 & 5.032 & 0.225 & \textbf{91.321} & 0.148 & 0.148 & \textbf{0.182} \\
& gated future-adapter & \textbf{3.462} & \textbf{4.453} & \textbf{0.140} & 100.280 & \textbf{0.282} & \textbf{0.282} & 0.184 \\
\midrule
OOD
& base pooled future-conditioned & 5.243 & 4.822 & 0.104 & \textbf{102.436} & 0.416 & 0.396 & \textbf{0.180} \\
& gated future-adapter & \textbf{3.002} & \textbf{2.698} & \textbf{0.057} & 105.067 & \textbf{0.642} & \textbf{0.646} & 0.211 \\
\bottomrule
\end{tabular}}
\end{table}

\subsubsection{Behavior under unreliable predictor futures}

We also test a key failure mode: biased futures could pull the UAV toward the wrong corridor. With an 8 m inference-time spatial bias, Table~\ref{tab:gfa_predictor_stress} shows that the adapter-only planner changes ADE only slightly but can lose safe-tracking time, whereas enabling RSEPSS keeps collision rate essentially unchanged and STT@8 degradation small ($-0.004$ ID, $-0.012$ OOD). This supports treating futures as gated residual guidance while leaving near-horizon safety to prefix repair. Additional ablations are in Appendix~\ref{app:ablation}.

\begin{table}[t]
\centering
\caption{Effect of unreliable predictor futures. Values report metric changes from clean predictor futures to an 8 m spatially biased predictor.}
\label{tab:gfa_predictor_stress}
\scriptsize
\setlength{\tabcolsep}{3.0pt}
\renewcommand{\arraystretch}{1.05}
\resizebox{\linewidth}{!}{
\begin{tabular}{llccccccc}
\toprule
\textbf{Split} & \textbf{Variant} & $\Delta$\textbf{ADE} & $\Delta$\textbf{FDE} & $\Delta$\textbf{CR} & $\Delta$\textbf{Smooth.} & $\Delta$\textbf{STT@8} & $\Delta$\textbf{STR@8} & $\Delta$\textbf{Time} \\
\midrule
ID  & gated future-adapter          & +0.022 & -0.157 & +0.011 & +5.396 & -0.061 & -0.061 & -0.064 \\
ID  & gated future-adapter + RSEPSS & +0.058 & -0.106 & +0.000 & +1.518 & -0.004 & -0.027 & -0.072 \\
OOD & gated future-adapter          & +0.069 & -0.017 & +0.008 & +0.918 & -0.042 & -0.042 & +0.006 \\
OOD & gated future-adapter + RSEPSS & +0.147 & +0.118 & -0.001 & -0.522 & -0.012 & -0.021 & +0.005 \\
\bottomrule
\end{tabular}}
\end{table}

\section{Conclusion and Limitations}
\label{sec:conclusion}

We introduced Future-Aware Flow Planning to address a gap in state-informed UAV target following: a follower must anticipate where the target is going, yet using predicted target paths as hard UAV references can make the near-horizon motion brittle or unsafe. Our approach separates what is predicted from what is executed. Target futures provide bounded, horizon-aligned residual guidance for clean UAV trajectory generation, while RSEPSS repairs only the executable prefix inside the sampling process. This turns future information into a planning cue rather than a commanded path, reducing late reactive following while preserving a receding-horizon safety interface. The ID/OOD benchmarks, ablations, and simulator-facing validation support this design as a safety-constrained following trade-off, with Future-MPC serving as a complementary hand-designed reference rather than an objective that one method universally dominates.

This study also has limitations. Simulator-facing stress tests in Appendix~\ref{app:sim_stress} examine interface perturbations and Isaac-style sensing, and the PX4 SITL validation in Section~\ref{sec:simulator_validation} still does not replace onboard perception or physical flight validation. The framework still assumes discrete target-state observations and local obstacle inputs, and does not cover end-to-end visual detection, re-identification, dynamic obstacles, or hardware-specific aerodynamics. RSEPSS improves executable safety but increases replanning cost, motivating lighter repair, distillation, or adaptive triggering for onboard deployment.

\acknowledgments{This work was supported by the Shenzhen Science and Technology Innovation Commission through the Major Science and Technology Project for Innovation and Entrepreneurship (Grant No. Z25306103, 2024). Computational resources were provided by the High-performance Computing Platform of Peking University.}

\bibliography{example}  

\appendix

\section{Evaluation Protocol and Metrics}
\label{app:setup}

\subsection{Benchmark, Baselines, and Simulator-Facing Protocol}
\label{app:benchmark_protocol}

We evaluate online receding-horizon UAV target following in cluttered 3D point-cloud environments. The ID benchmark uses forest scenes with low, medium, and high obstacle densities. The OOD benchmark introduces controlled shifts in obstacle shape, obstacle distribution, target speed, and their combination, following common simulation generalization practice where geometry, procedural factors, or dynamics are varied beyond nominal training conditions~\citep{tobin2017domain,peng2018sim,cobbe2020procgen}. All methods use the same saved fixed cases and rollout protocol: 150 effective ID episodes (50 low-density, 50 medium-density, and 50 high-density after one skipped target-path generation case) and 200 OOD episodes (50 per shift).

The main comparison includes adapted SafeFlow~\citep{dai2025safeflow} and SafeFlowMatcher~\citep{yang2025safeflowmatcher} baselines under the same predictor, planner input, sampling budget, obstacle budget, execution prefix, replanning limit, and metrics. We also include Future-MPC as a fixed hand-designed receding-horizon reference that uses the same target predictor and rollout protocol but replaces learned trajectory generation with its native objective and cylinder-geometry clearance. The main tables report tracking accuracy, collision rate, jerk-RMS smoothness, safe tracking time and success at an 8 m threshold (STT@8/STR@8), and seconds per replan.

We additionally run method-only simulator-facing stress tests in the appendix. L1 perturbs target-state, point-cloud, and latency interfaces, and L2 uses Isaac-style LiDAR-derived local point clouds under nominal and degraded sensing profiles. Compact summaries are reported in Fig.~\ref{fig:sim_stress_bars}, with protocol details and visualization in Appendix~\ref{app:sim_stress}. The PX4 flight-controller-in-the-loop validation is reported separately in Section~\ref{sec:simulator_validation}, with uncertainty in Appendix~\ref{app:statistical_stability}.

\subsection{Metric Definitions}
\label{app:metrics_def}

All metrics are computed from the same fixed receding-horizon rollout protocol used in the main tables. For episode $i$, let $p_{i,t}^{\mathrm{uav}}$ and $p_{i,t}^{\mathrm{tar}}$ denote the executed UAV and target positions at rollout time $t\in\{1,\ldots,T_i\}$. Let $h_{i,t}\in\{0,1\}$ be the dense hard-collision indicator at that executed state, and let $d_{i,t}=\|p_{i,t}^{\mathrm{uav}}-p_{i,t}^{\mathrm{tar}}\|_2$ be the tracking distance.

\paragraph{Tracking error.}
Average displacement error and final displacement error are defined as
\begin{equation}
\mathrm{ADE}=\frac{1}{N}\sum_{i=1}^{N}\frac{1}{T_i}\sum_{t=1}^{T_i}d_{i,t},
\qquad
\mathrm{FDE}=\frac{1}{N}\sum_{i=1}^{N}d_{i,T_i}.
\end{equation}
ADE measures rollout-level tracking accuracy, whereas FDE emphasizes the terminal tracking state after repeated replanning.

\paragraph{Collision rate.}
The collision rate reported in the main tables is the mean dense hard-collision rate over executed rollout states:
\begin{equation}
\mathrm{CR}=\frac{1}{N}\sum_{i=1}^{N}\frac{1}{T_i}\sum_{t=1}^{T_i}h_{i,t}.
\end{equation}
This state-level definition is more informative for receding-horizon tracking than a single episode-level collision flag, because it captures how often the executed prefix enters collision across the closed-loop rollout.

\paragraph{Trajectory smoothness.}
Let $\Delta t$ be the simulation time step. We compute velocity, acceleration, and jerk by finite differences,
\begin{equation}
v_{i,t}=\frac{p_{i,t}-p_{i,t-1}}{\Delta t},\quad
 a_{i,t}=\frac{v_{i,t}-v_{i,t-1}}{\Delta t},\quad
 j_{i,t}=\frac{a_{i,t}-a_{i,t-1}}{\Delta t}.
\end{equation}
The smoothness score is jerk RMS,
\begin{equation}
\mathrm{Smooth}=\frac{1}{N}\sum_{i=1}^{N}
\sqrt{\frac{1}{T_i-3}\sum_{t=4}^{T_i}\|j_{i,t}\|_2^2}.
\end{equation}
Lower values indicate smoother executed trajectories.

\paragraph{Safe tracking.}
For a tracking tolerance $r$, define the safe-tracking indicator
\begin{equation}
s_{i,t}^{(r)}=(1-h_{i,t})\,\mathbf{1}\{d_{i,t}\le r\}.
\end{equation}
Safe tracking time is the mean fraction of rollout time satisfying this criterion,
\begin{equation}
\mathrm{STT@}r=\frac{1}{N}\sum_{i=1}^{N}\frac{1}{T_i}\sum_{t=1}^{T_i}s_{i,t}^{(r)}.
\end{equation}
Safe tracking success rate is the episode-level rate of remaining collision-free and ending within the same tolerance,
\begin{equation}
\mathrm{STR@}r=\frac{1}{N}\sum_{i=1}^{N}\mathbf{1}\left\{\sum_{t=1}^{T_i}h_{i,t}=0\right\}\mathbf{1}\{d_{i,T_i}\le r\}.
\end{equation}
The main paper reports $r=8\,\mathrm{m}$.

\paragraph{Online replanning time.}
Runtime is reported as seconds per replanning call,
\begin{equation}
\mathrm{Time}=\frac{1}{N}\sum_{i=1}^{N}\frac{\mathrm{wall\_time}_i}{R_i},
\end{equation}
where $R_i$ is the number of replanning calls in episode $i$. Timing values are compared only within the same fixed evaluation runner and rollout protocol.

\subsection{Rollout Protocol}
\label{app:rollout_protocol}

All main-table and paper-facing ablation results use the same fixed benchmark cases and online rollout settings: horizon $H=64$, target-history length $L=16$, executed prefix length $K_{\mathrm{exec}}=4$, observation interval of 4 simulation steps, 12 flow sampling updates per replan, $\Delta t=0.2\,\mathrm{s}$, local obstacle radius $R_{\mathrm{obs}}=8.0\,\mathrm{m}$, obstacle sample count $\mathrm{obs\_sample\_M}=256$, fixed noise seed 0, episode seed 123, and maximum replanning count $25$. ID evaluation is grouped by obstacle density (low, medium, high). OOD evaluation uses the four controlled shifts described below. The reported main-table entries are fixed-case means; statistical stability is assessed by the bootstrap protocol in Appendix~\ref{app:statistical_stability}.

\paragraph{Baseline conditioning and controlled variables.}
The SafeFlow and SafeFlowMatcher rows are controlled adaptations, not claims of reproducing every system-level detail of the original papers. They use the same target-motion predictor, base learned x-pred planner family, candidate budget, obstacle samples, execution prefix, replanning limit, and metric implementation as our method; the isolated difference is the safety-aware sampling or correction mechanism. Future-MPC uses the same target predictor and fixed online rollout protocol but replaces learned trajectory generation with a receding-horizon target-following objective and local cylinder-geometry clearance. We include it as a fixed hand-designed receding-horizon reference, not as a learned generative-planner ablation. In the gated future-adapter ablation, the base pooled future-conditioned control keeps the predictor and base planner condition unchanged and disables only the auxiliary residual future-adapter pathway. In the RSEPSS ablation, the gated future-adapter planner is held fixed and only embedded executable-prefix repair is toggled. In the x-pred versus v-pred diagnostic, both variants use the same predictor conditioning and no RSEPSS.

\subsection{Statistical Stability}
\label{app:statistical_stability}

For fixed-case benchmark tables, uncertainty is estimated by non-parametric bootstrap over evaluation episodes. ID resampling is stratified by obstacle-density group, and OOD resampling is stratified by shift type, preserving the macro-averaging structure used in Tables~\ref{tab:id_main}--\ref{tab:ood_main}. We use 10,000 bootstrap resamples. Table~\ref{tab:main_bootstrap_ci} reports compact uncertainty for the claim-critical safety--tracking metrics in the two main tables: FDE, collision rate, STT@8, and STR@8. For the high-fidelity PX4 flight-controller-in-the-loop subset, all paper-facing statistics are computed over 120 completed runs: sixty ID forest-medium scenes and sixty OOD combined-shift wall scenes. Table~\ref{tab:px4_stat_stability} reports the corresponding grouped means, standard errors, and bootstrap intervals.

\begin{table}[t]
\centering
\caption{Bootstrap uncertainty for the claim-critical metrics in Tables~\ref{tab:id_main}--\ref{tab:ood_main}. Each cell reports mean (SE) [95\% bootstrap CI]; resampling preserves the ID density and OOD shift macro-averaging.}
\label{tab:main_bootstrap_ci}
\scriptsize
\setlength{\tabcolsep}{1.6pt}
\renewcommand{\arraystretch}{1.05}
\resizebox{\linewidth}{!}{
\begin{tabular}{llcccc}
\toprule
\textbf{Split} & \textbf{Method} & \textbf{FDE}$\downarrow$ & \textbf{CR}$\downarrow$ & \textbf{STT@8}$\uparrow$ & \textbf{STR@8}$\uparrow$ \\
\midrule
ID & SafeFlow & 5.622 (0.077) [5.482, 5.785] & 0.006 (0.002) [0.002, 0.011] & 0.938 (0.018) [0.900, 0.972] & 0.933 (0.021) [0.892, 0.973] \\
ID & SafeFlowMatcher & 8.747 (0.103) [8.545, 8.950] & 0.000 (0.000) [0.000, 0.000] & 0.740 (0.018) [0.704, 0.775] & 0.469 (0.041) [0.389, 0.550] \\
ID & Future-MPC & 3.534 (0.095) [3.348, 3.719] & 0.043 (0.005) [0.033, 0.053] & 0.583 (0.040) [0.503, 0.664] & 0.583 (0.040) [0.509, 0.664] \\
ID & Ours & 5.448 (0.119) [5.222, 5.688] & 0.000 (0.000) [0.000, 0.000] & 0.961 (0.009) [0.942, 0.977] & 0.913 (0.023) [0.865, 0.953] \\
\midrule
OOD & SafeFlow & 3.927 (0.123) [3.687, 4.165] & 0.014 (0.005) [0.006, 0.024] & 0.814 (0.046) [0.723, 0.900] & 0.812 (0.048) [0.708, 0.896] \\
OOD & SafeFlowMatcher & 7.177 (0.288) [6.614, 7.747] & 0.009 (0.004) [0.002, 0.019] & 0.733 (0.042) [0.647, 0.813] & 0.562 (0.069) [0.417, 0.688] \\
OOD & Future-MPC & 4.288 (0.151) [3.982, 4.579] & 0.003 (0.002) [0.000, 0.008] & 0.958 (0.028) [0.896, 1.000] & 0.958 (0.028) [0.896, 1.000] \\
OOD & Ours & 3.719 (0.366) [3.080, 4.487] & 0.002 (0.001) [0.000, 0.005] & 0.896 (0.031) [0.831, 0.950] & 0.854 (0.041) [0.771, 0.938] \\
\bottomrule
\end{tabular}}
\end{table}

\begin{table}[t]
\centering
\caption{Statistical stability for the PX4 flight-controller-in-the-loop validation. Each cell reports mean (SE) [95\% bootstrap CI] over scenes. Target metrics are path-aligned to the reference-execution phase.}
\label{tab:px4_stat_stability}
\scriptsize
\setlength{\tabcolsep}{2.0pt}
\renewcommand{\arraystretch}{1.05}
\resizebox{\linewidth}{!}{
\begin{tabular}{lccccc}
\toprule
\textbf{Split} & \textbf{Ref. RMSE}$\downarrow$ & \textbf{Target RMSE}$\downarrow$ & \textbf{CR}$\downarrow$ & \textbf{STT@8}$\uparrow$ & \textbf{STR@8}$\uparrow$ \\
\midrule
ID, $n=60$  & 1.84 (0.20) [1.49, 2.19] & 5.12 (0.63) [3.99, 6.20] & 0.00 (0.00) [0.00, 0.00] & 0.95 (0.05) [0.86, 1.00] & 0.83 (0.17) [0.50, 1.00] \\
OOD, $n=60$ & 2.02 (0.22) [1.60, 2.39] & 5.40 (1.18) [3.42, 7.54] & 0.00 (0.00) [0.00, 0.00] & 0.82 (0.12) [0.60, 1.00] & 0.67 (0.21) [0.33, 1.00] \\
All, $n=120$ & 1.93 (0.14) [1.65, 2.19] & 5.26 (0.64) [4.10, 6.52] & 0.00 (0.00) [0.00, 0.00] & 0.89 (0.06) [0.75, 1.00] & 0.75 (0.13) [0.50, 1.00] \\
\bottomrule
\end{tabular}}
\end{table}

\section{Implementation and Training Details}
\label{app:implementation_training}

All learned planners use a Transformer-style trajectory backbone with $d_{\mathrm{model}}=256$, 6 layers, 8 attention heads, an obstacle-token cap of 128, and horizon length 64. The target-motion predictor is a Mamba-based forecaster with history length 16, horizon length 64, and 4 selected future hypotheses. The aligned training records are successful expert samples from scene-aligned JSONL files; scenes are split by scene id with seed 0 into 80\% training, 10\% validation, and 10\% test partitions. The expert trajectories in the aligned dataset are generated by four OMPL-style geometric planners (RRTstar, RRTConnect, BITstar, and PRMstar), and only records marked successful are used for model training and evaluation. The predictor is trained for 15 epochs with AdamW, learning rate $2\times10^{-4}$, weight decay $10^{-4}$, and gradient clipping at 1.0. The clean-trajectory base planner is trained with the direct $x_1$ prediction objective described in Section~\ref{sec:flow_planner}; the final gated adapter is then tuned with the predictor and base branch fixed, so that the adapter learns a conservative residual refinement rather than replacing the base generator.

For the final adapter run, we use 200 scene-aligned training files, batch size 8, AdamW with learning rate $2\times10^{-4}$, weight decay $10^{-4}$, gradient clipping at 1.0, and 25 epochs; model selection uses the validation score, with the reported run selecting epoch 5. The adapter uses the last planner block, bottleneck size 64, maximum residual scale 0.015, future-dropout probability 0.05, and a prefix-adapter length of 4 to match the executed prefix. The fixed-evaluation runner uses a fixed noise seed, 12 flow sampling updates per replan, and the rollout parameters in Appendix~\ref{app:rollout_protocol}. RSEPSS is embedded in the inference-time flow sampling loop: each clean candidate is prefix-scored and locally repaired before being used in the next flow update. It does not introduce additional trained parameters or backpropagated losses. Its final correction configuration uses $K_{\mathrm{corr}}=8$, safety margin $d_{\mathrm{safe}}=0.60\,\mathrm{m}$, trigger margin $d_{\mathrm{trigger}}=0.75\,\mathrm{m}$, top-3 risky prefix points, 3 projection iterations, one smoothing iteration with smoothing weight 0.05, world-step cap $0.80\,\mathrm{m}$, and start-point anchoring.

\section{Additional Predictor and Benchmark Results}
\label{app:additional_results}

\subsection{Target Predictor Quality Probe}
\label{app:predictor_quality}

Because the planner conditions on predicted target futures, we separately evaluate the target-motion predictor before testing the downstream planner. The open-loop probe uses the same Mamba predictor checkpoint as the planner evaluations and samples 160 held-out records from each target-motion complexity bin. Metrics are computed between the predicted target future and the ground-truth target future, not between the UAV and the target.

\begin{table}[t]
\centering
\caption{Open-loop target predictor quality on held-out records. Complexity is the trajectory-complexity score used for binning; finite rate reports the fraction of numerically valid predicted futures.}
\label{tab:app_predictor_quality}
\scriptsize
\setlength{\tabcolsep}{4.0pt}
\renewcommand{\arraystretch}{1.05}
\begin{tabular}{lccccc}
\toprule
\textbf{Motion bin} & \textbf{Records} & \textbf{Complexity} & \textbf{Target ADE}$\downarrow$ & \textbf{Target FDE}$\downarrow$ & \textbf{Finite}$\uparrow$ \\
\midrule
Simple  & 160 & 0.027 & 0.213 & 0.448 & 1.000 \\
Complex & 160 & 0.724 & 0.278 & 0.640 & 1.000 \\
\bottomrule
\end{tabular}
\end{table}

The predictor is not treated as an oracle: its outputs are passed to the planner as uncertain feature guidance, and the biased-future stress test in Table~\ref{tab:gfa_predictor_stress} explicitly measures what happens when the predicted futures are wrong. Additional fixed-case ablations show that the predictor improves target-following accuracy when evaluated under a matched rollout protocol. Together with the biased-future stress test, this suggests that predicted target futures are useful guidance but should not be treated as hard trajectory anchors.

\subsection{Detailed Main Comparison Tables}
\label{app:detailed_main_tables}

Tables~\ref{tab:id_main_full} and~\ref{tab:ood_main_full} provide the full group-wise results summarized in Tables~\ref{tab:id_main} and~\ref{tab:ood_main}.

\begin{table*}[t]
\centering
\caption{Full ID comparison by forest density. Smoothness is measured by jerk RMS.}
\label{tab:id_main_full}
\scriptsize
\setlength{\tabcolsep}{3.2pt}
\renewcommand{\arraystretch}{1.05}
\resizebox{\textwidth}{!}{
\begin{tabular}{llccccccc}
\toprule
\textbf{Density} & \textbf{Method} & \textbf{ADE}$\downarrow$ & \textbf{FDE}$\downarrow$ & \textbf{CR}$\downarrow$ & \textbf{Smooth.}$\downarrow$ & \textbf{STT@8}$\uparrow$ & \textbf{STR@8}$\uparrow$ & \textbf{Time}$\downarrow$ \\
\midrule
Low
& SafeFlow        & 4.181 & 5.570 & 0.001 & 104.953 & 0.968 & 0.960 & 0.211 \\
& SafeFlowMatcher & 5.999 & 8.706 & \textbf{0.000} & 115.929 & 0.764 & 0.500 & \textbf{0.046} \\
& Future-MPC      & \textbf{3.976} & \textbf{3.458} & 0.045 & \textbf{27.410} & 0.600 & 0.600 & 0.123 \\
& Ours            & 4.258 & 5.228 & \textbf{0.000} & 82.352 & \textbf{0.987} & \textbf{0.980} & 0.283 \\
\midrule
Medium
& SafeFlow        & 4.187 & 5.655 & 0.006 & 104.036 & 0.937 & \textbf{0.940} & 0.216 \\
& SafeFlowMatcher & 6.105 & 8.578 & \textbf{0.000} & 114.038 & 0.733 & 0.500 & \textbf{0.046} \\
& Future-MPC      & \textbf{3.898} & \textbf{3.432} & 0.045 & \textbf{27.909} & 0.640 & 0.640 & 0.126 \\
& Ours            & 4.490 & 5.601 & \textbf{0.000} & 82.626 & \textbf{0.947} & 0.880 & 0.310 \\
\midrule
High
& SafeFlow        & 4.015 & 5.641 & 0.010 & 100.359 & 0.910 & \textbf{0.898} & 0.213 \\
& SafeFlowMatcher & 6.021 & 8.957 & \textbf{0.000} & 111.378 & 0.723 & 0.408 & \textbf{0.047} \\
& Future-MPC      & \textbf{3.769} & \textbf{3.713} & 0.038 & \textbf{27.345} & 0.510 & 0.510 & 0.125 \\
& Ours            & 4.348 & 5.514 & \textbf{0.000} & 81.305 & \textbf{0.947} & 0.878 & 0.307 \\
\bottomrule
\end{tabular}}
\end{table*}

\begin{table*}[t]
\centering
\caption{Full OOD comparison by shift type. Smoothness is measured by jerk RMS.}
\label{tab:ood_main_full}
\scriptsize
\setlength{\tabcolsep}{3.2pt}
\renewcommand{\arraystretch}{1.05}
\resizebox{\textwidth}{!}{
\begin{tabular}{llccccccc}
\toprule
\textbf{Shift} & \textbf{Method} & \textbf{ADE}$\downarrow$ & \textbf{FDE}$\downarrow$ & \textbf{CR}$\downarrow$ & \textbf{Smooth.}$\downarrow$ & \textbf{STT@8}$\uparrow$ & \textbf{STR@8}$\uparrow$ & \textbf{Time}$\downarrow$ \\
\midrule
Shape
& SafeFlow        & 3.057 & 3.766 & \textbf{0.000} & 102.190 & \textbf{1.000} & \textbf{1.000} & 0.196 \\
& SafeFlowMatcher & 4.903 & 7.603 & 0.011 & 122.214 & 0.845 & 0.750 & \textbf{0.043} \\
& Future-MPC      & 5.205 & 4.958 & \textbf{0.000} & \textbf{27.445} & \textbf{1.000} & \textbf{1.000} & 0.122 \\
& Ours            & \textbf{2.877} & \textbf{3.058} & \textbf{0.000} & 87.985 & \textbf{1.000} & \textbf{1.000} & 0.221 \\
\midrule
Distribution
& SafeFlow        & 2.957 & 4.141 & 0.015 & 103.912 & 0.750 & 0.750 & 0.204 \\
& SafeFlowMatcher & 4.948 & 7.855 & 0.008 & 117.573 & 0.795 & 0.583 & \textbf{0.044} \\
& Future-MPC      & 4.953 & 4.306 & \textbf{0.004} & \textbf{27.323} & \textbf{0.917} & \textbf{0.917} & 0.124 \\
& Ours            & \textbf{2.834} & \textbf{3.536} & \textbf{0.004} & 86.270 & \textbf{0.917} & \textbf{0.917} & 0.301 \\
\midrule
Speed
& SafeFlow        & 3.623 & 3.450 & \textbf{0.000} & 113.829 & 0.984 & \textbf{1.000} & 0.182 \\
& SafeFlowMatcher & 5.317 & 5.030 & \textbf{0.000} & 125.800 & 0.807 & 0.583 & \textbf{0.041} \\
& Future-MPC      & 4.366 & 3.726 & \textbf{0.000} & \textbf{27.003} & \textbf{1.000} & \textbf{1.000} & 0.123 \\
& Ours            & \textbf{3.464} & \textbf{2.221} & \textbf{0.000} & 98.288 & 0.975 & \textbf{1.000} & 0.220 \\
\midrule
Combined
& SafeFlow        & \textbf{4.434} & 4.349 & 0.042 & 120.145 & 0.521 & 0.500 & 0.215 \\
& SafeFlowMatcher & 6.935 & 8.219 & 0.019 & 121.482 & 0.485 & 0.333 & \textbf{0.044} \\
& Future-MPC      & 4.555 & \textbf{4.164} & 0.008 & \textbf{27.135} & \textbf{0.917} & \textbf{0.917} & 0.125 \\
& Ours            & 5.438 & 6.059 & \textbf{0.004} & 86.777 & 0.692 & 0.500 & 0.405 \\
\bottomrule
\end{tabular}}
\end{table*}

\section{OOD Benchmark Design and Qualitative Results}
\label{app:ood_qualitative}
\label{app:ood_settings}

The OOD benchmark is designed to separate different deployment-time mismatches rather than reporting a single aggregated robustness number. This follows the same principle as domain and dynamics randomization in robotic simulation and procedurally generated generalization benchmarks: robustness should be tested by controlled changes along interpretable axes, not only by evaluating on another random split~\citep{tobin2017domain,peng2018sim,cobbe2020procgen}. We therefore use four shift types. \emph{Shape shift} changes obstacle geometry and layout while preserving the online tracking interface. \emph{Distribution shift} changes the statistics and spatial arrangement of obstacles, which alters the local point observations encountered during rollout. \emph{Speed shift} changes target-motion dynamics and directly stresses future target anticipation. \emph{Combined shift} applies multiple changes together and serves as the hardest OOD setting.

\begin{figure}[t]
    \centering
    \includegraphics[width=0.72\linewidth]{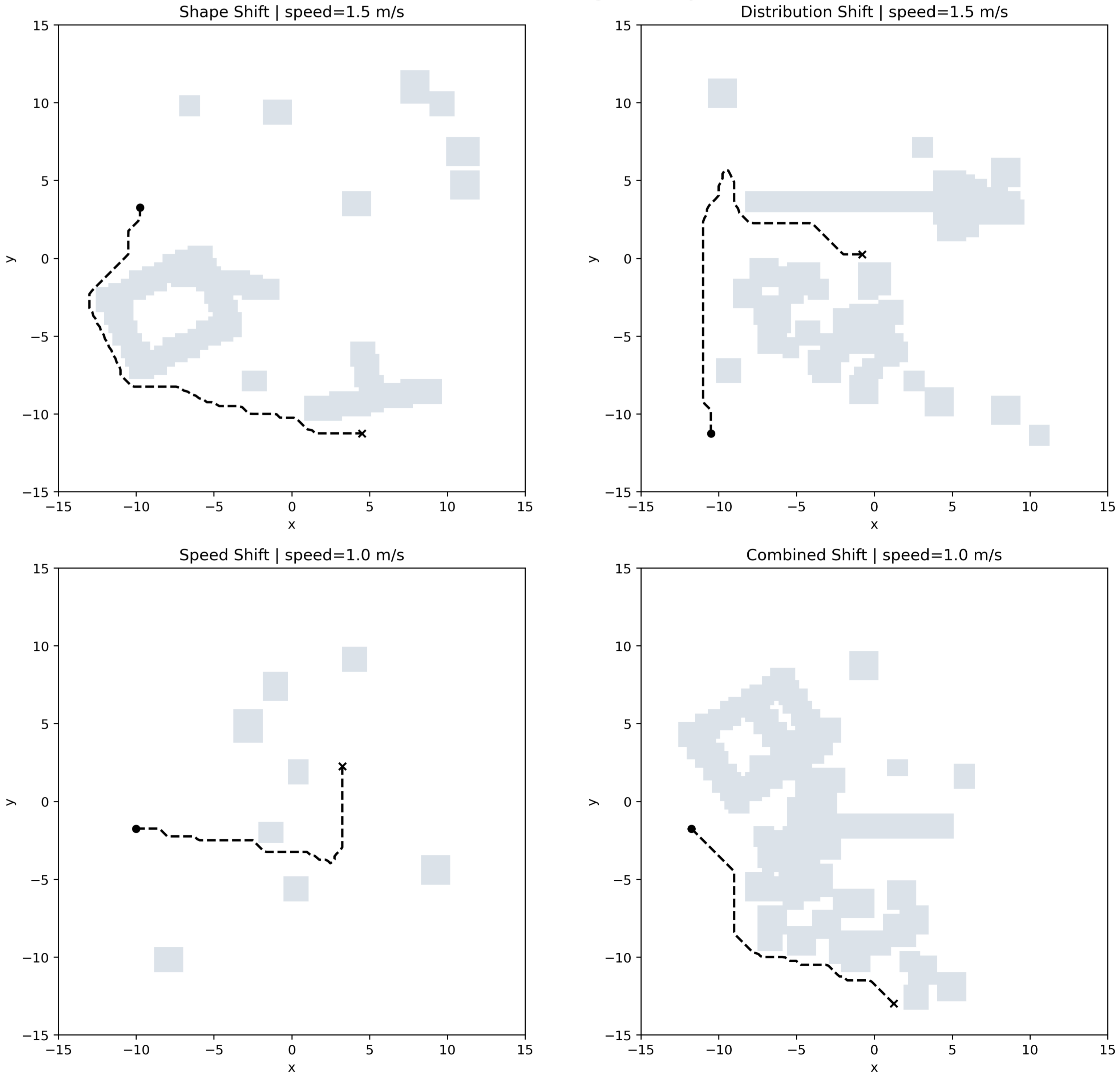}
    \caption{OOD scene design. The benchmark separates obstacle-shape, obstacle-distribution, target-speed, and combined shifts to test different forms of deployment mismatch.}
    \label{fig:ood_scene_gallery}
\end{figure}

Figure~\ref{fig:ood_scene_gallery} visualizes the scenario-level design. The shape and distribution shifts mainly perturb the obstacle-related condition received by the planner, whereas the speed shift perturbs the target-future condition used by the predictor and future adapter. The combined setting couples these two sources of mismatch, making it useful for testing whether a method can preserve both tracking and executable safety when scene geometry and target dynamics depart from the ID distribution.

\begin{figure}[t]
    \centering
    \includegraphics[width=0.95\linewidth]{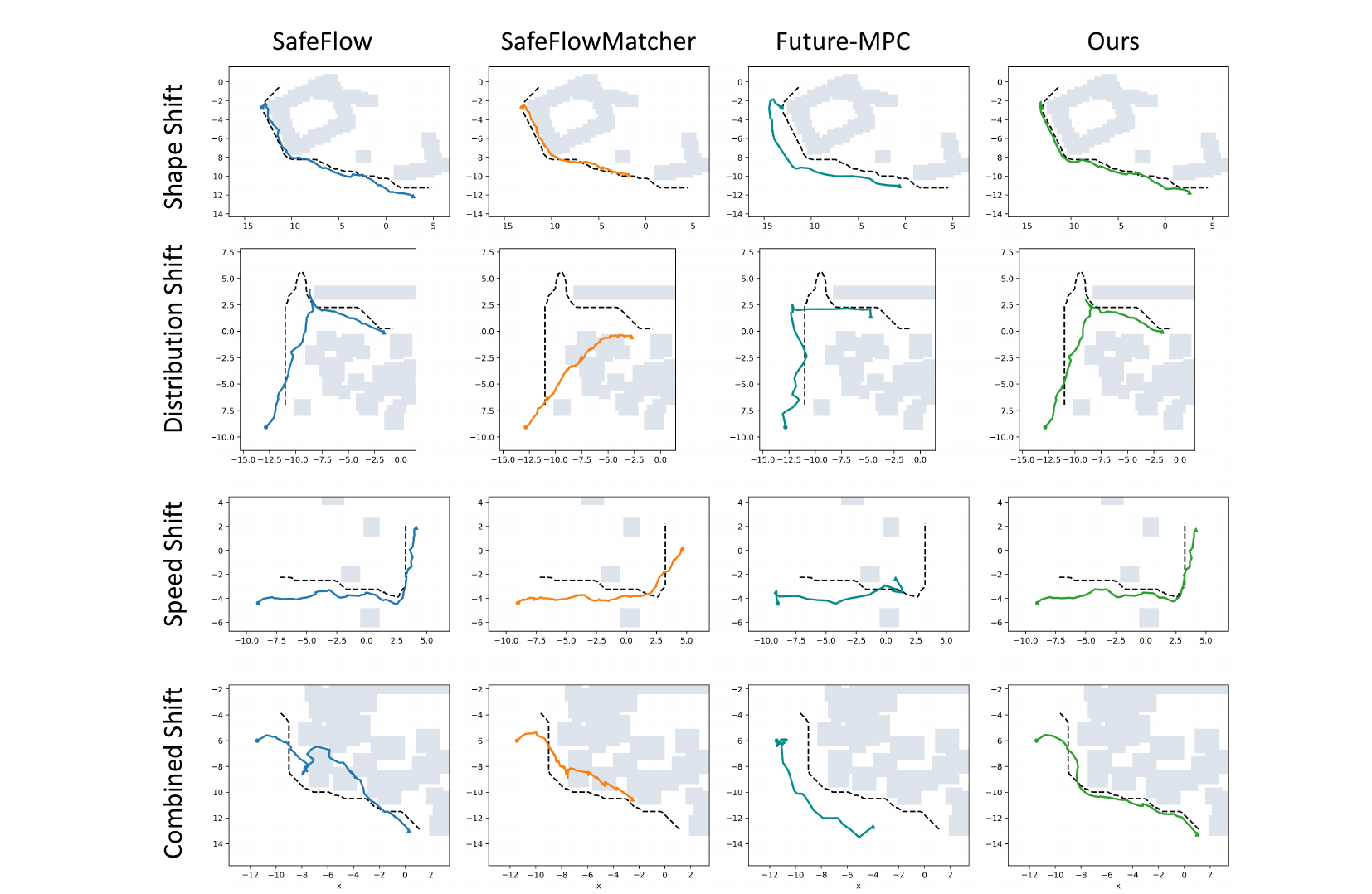}
    \caption{OOD qualitative trajectory comparison across the four displayed methods. The visualization complements Table~\ref{tab:ood_main} by showing how safety-aware flow baselines, Future-MPC, and the proposed method trade target tracking, clearance, and executable smoothness under shifted scenes.}
    \label{fig:ood_rollout_qual}
\end{figure}

Figure~\ref{fig:ood_rollout_qual} uses a common 2D rollout visualization protocol for the four displayed methods: each OOD row uses the same selected scene, local obstacle rendering, target trajectory, axis bounds, and method ordering so that qualitative differences can be compared directly. The OOD rollout visualization is intended to show the failure modes behind Table~\ref{tab:ood_main}. SafeFlow improves safety through stronger sampling-time correction, but can become more conservative. SafeFlowMatcher is computationally light yet more prone to weakened tracking in difficult shifts. Future-MPC serves as a fixed optimization-based reference, while the proposed method emphasizes learned generation with local prefix safety in several shifted scenes.

\section{Simulator-Facing Stress Tests}
\label{app:sim_stress}

\begin{figure}[t]
\centering
\includegraphics[width=0.95\linewidth]{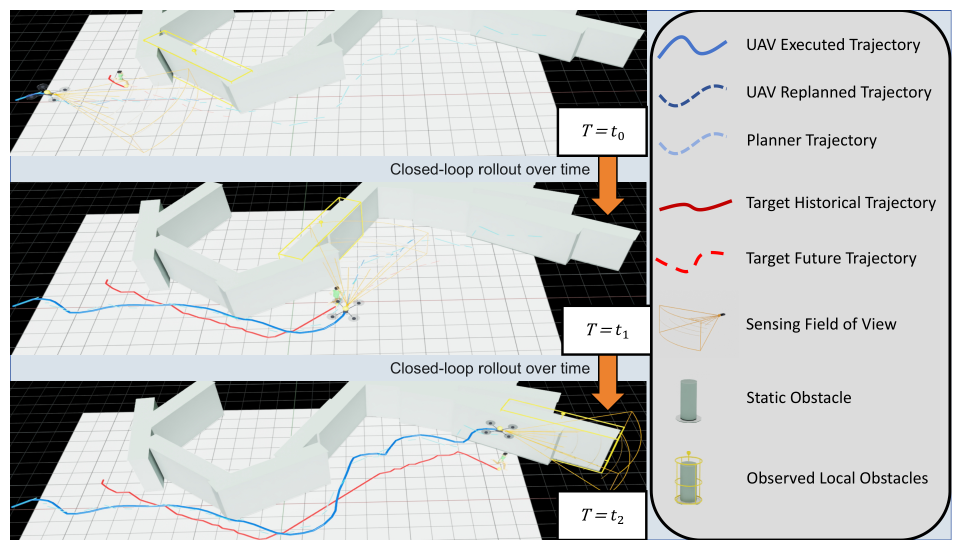}
\caption{Representative simulator-facing visualization in Isaac Sim. The sequence shows closed-loop rollout over time, including the executed UAV trajectory, replanned and planner-reference trajectories, target trajectory, sensing field of view, static obstacles, and locally observed obstacles.}
\label{fig:app_sim_exp}
\end{figure}

The main benchmark isolates planner quality under a fixed receding-horizon rollout protocol. To further probe deployment-facing robustness of the final method, we run two method-only stress tests in the appendix. These tests are not used as an additional baseline comparison; they examine whether the final planner remains stable when its target-state interface, obstacle-observation interface, and sensing profile are made less ideal. Figure~\ref{fig:app_sim_exp} shows the corresponding Isaac Sim visualization elements for a representative rollout. The higher-fidelity PX4 flight-controller-in-the-loop validation is kept in the main paper to avoid mixing execution-level evidence with the lighter interface and sensing stress plots.

\begin{figure}[t]
\centering
\begin{subfigure}[t]{0.43\linewidth}
    \centering
    \includegraphics[width=\linewidth]{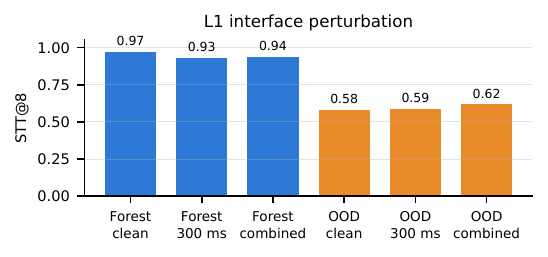}
    \caption{L1}
    \label{fig:sim_l1_bars}
\end{subfigure}
\hfill
\begin{subfigure}[t]{0.43\linewidth}
    \centering
    \includegraphics[width=\linewidth]{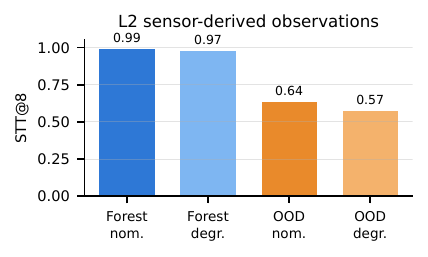}
    \caption{L2}
    \label{fig:sim_l2_bars}
\end{subfigure}
\caption{Simulator-facing stress tests for the final method. L1 reports STT@8 under selected interface perturbations; L2 reports STT@8 under nominal and degraded sensor-derived observations.}
\label{fig:sim_stress_bars}
\end{figure}

\paragraph{L1: interface perturbations.}
L1 keeps the rollout protocol unchanged but perturbs the planner inputs at the interface level. We perturb target states, degrade obstacle point-cloud observations, add observation latency up to 300 ms, and evaluate a combined mild stress profile. Across 682 records, the forest/medium setting remains collision-free under all L1 stresses, and STT@8 stays above 0.93 even under 300 ms latency and combined mild stress (Fig.~\ref{fig:sim_l1_bars}). In the harder OOD combined-shift setting, performance degrades as expected, but the method still preserves non-zero safe tracking under all tested perturbation profiles.

\paragraph{L2: Isaac-style sensing.}
L2 replaces the analytic local obstacle input with a LiDAR-style local point cloud generated from nominal and degraded sensing profiles. The degraded profile shortens range from 8.0 m to 6.5 m, narrows horizontal field of view from 90 degrees to 70 degrees, reduces angular beams by 37.5\%, increases dropout from 0.05 to 0.22, and increases target-observation noise. Under this sensor-derived observation chain, the method preserves the forest trend, while the OOD combined-shift setting remains the most difficult case (Fig.~\ref{fig:sim_l2_bars}).

\section{Robustness Analysis under OOD Shifts}
\label{app:robustness}
\label{app:theory}

This section gives a conditional robustness analysis for the OOD results. The goal is not to claim a universal safety certificate. Instead, we make explicit how the three planner-side choices used in the method, namely reliability-gated future adaptation, clean-trajectory x-pred generation, and embedded executable-prefix repair, reduce distinct terms in an OOD risk decomposition under standard boundedness and Lipschitz assumptions.

\subsection{Notation and Assumptions}

Let $c$ denote the observed planning condition, including UAV state, target history, and local obstacle observations. Let $Z$ be the predicted future-token representation produced from the target predictor, and let $Z^\star$ denote the ideal future-token representation induced by the ground-truth future target motion. The clean trajectory predicted by the planner before safety repair is written as
\begin{equation}
\hat{x}_1=\Pi_\theta(c,Z),
\end{equation}
and the embedded RSEPSS repair operator with margin $m$ is written as $S_m$. The executed candidate is therefore
\begin{equation}
\tau=S_m(\hat{x}_1)=S_m\bigl(\Pi_\theta(c,Z)\bigr).
\end{equation}
For analysis, let $\ell(\tau,c)$ be a task loss that can include tracking error, collision risk, smoothness, and executable tracking failure. We assume that $\ell$ is locally Lipschitz in the trajectory and condition in the neighborhood of the evaluated rollouts. This assumption is standard for local robustness statements and is weaker than requiring global robustness over arbitrary scenes.

\subsection{Bounded Influence of Unreliable Future Tokens}

The future adapter is designed so that predicted futures influence the planner as residual feature guidance rather than as hard trajectory anchors. For a late adapted block $l$, write the future residual as
\begin{equation}
\Delta H_l(Z)=\alpha_l g_l(Z)R_l(H_l,Z),
\end{equation}
where $0\le g_l(Z)\le 1$ is the reliability gate, $\alpha_l$ is a small learnable scale, and $R_l$ is the cross-attention adapter output. Assume that, for fixed $H_l$, $R_l$ is Lipschitz in $Z$ with constant $L_R$, $g_l$ is Lipschitz in $Z$ with constant $L_g$, and $\|R_l(H_l,Z)\|\le B_R$. Then, for a corrupted future token $Z+\delta Z$,
\begin{align}
\|\Delta H_l(Z+\delta Z)-\Delta H_l(Z)\|
&=
\alpha_l\|g_l(Z+\delta Z)R_l(H_l,Z+\delta Z)-g_l(Z)R_l(H_l,Z)\| \\
&\le
\alpha_l\left(L_R+B_RL_g\right)\|\delta Z\|.
\end{align}
If the downstream decoder and intermediate blocks from the adapted layers to the clean trajectory head have Lipschitz constant $L_\Pi$, the induced clean-trajectory perturbation satisfies
\begin{equation}
\|\Pi_\theta(c,Z+\delta Z)-\Pi_\theta(c,Z)\|
\le
L_\Pi
\sum_{l\in\mathcal{A}}
\alpha_l\left(L_R+B_RL_g\right)\|\delta Z\|,
\label{eq:future_adapter_bound}
\end{equation}
where $\mathcal{A}$ is the set of adapted blocks. Equation~\ref{eq:future_adapter_bound} formalizes why the adapter is less brittle than coordinate-level future anchoring: the effect of predictor corruption is mediated by the residual scale, the gate, and the feature-level adapter, rather than being imposed directly on the generated UAV trajectory.

\subsection{Clean-Trajectory x-pred and Robust Task Alignment}

We next analyze why predicting the clean trajectory is aligned with robustness for this task. Let $x_1$ be a feasible expert trajectory and let $\mathcal{R}(x)$ be a task-level risk functional over clean future trajectories, such as tracking error, smoothness, or executable-safety loss. Assume that $\mathcal{R}$ is locally Lipschitz near feasible UAV tracking trajectories:
\begin{equation}
|\mathcal{R}(x)-\mathcal{R}(y)|
\le
L_{\mathcal R}\|x-y\|.
\end{equation}
For x-pred, the planner outputs $\hat{x}_1$ directly. Therefore,
\begin{equation}
|\mathcal{R}(\hat{x}_1)-\mathcal{R}(x_1)|
\le
L_{\mathcal R}\|\hat{x}_1-x_1\|.
\label{eq:xpred_task_risk}
\end{equation}
Thus, reducing the x-pred clean-trajectory error directly reduces task-risk deviation.

By contrast, a velocity-style parameterization predicts an auxiliary quantity $\hat{v}$ and recovers the clean trajectory through a reconstruction map. Under the linear flow interpolation $x_\lambda=(1-\lambda)x_0+\lambda x_1$, the velocity target is $v=x_1-x_0$ and the clean trajectory is reconstructed as
\begin{equation}
\hat{x}_1=\Psi_\lambda(x_\lambda,\hat{v})=x_\lambda+(1-\lambda)\hat{v}.
\end{equation}
Then
\begin{equation}
|\mathcal{R}(\Psi_\lambda(x_\lambda,\hat{v}))-\mathcal{R}(x_1)|
\le
L_{\mathcal R}(1-\lambda)\|\hat{v}-v\|.
\end{equation}
This bound is valid, but it controls task risk only through an auxiliary reconstruction. More importantly for our method, safety and tracking corrections are naturally defined in clean trajectory space. If a clean-space correction $\Delta x_1$ must be inserted during sampling, the equivalent change in the velocity parameterization is
\begin{equation}
\Delta v=\frac{\Delta x_1}{1-\lambda},
\end{equation}
which becomes ill-conditioned as $\lambda$ approaches the clean endpoint. In contrast, x-pred allows the corrected clean trajectory $S_m(\hat{x}_1)$ to be written back directly into the sampling update. This provides a mathematical reason for pairing embedded RSEPSS with a clean-trajectory prediction head.

The same conclusion can be stated in terms of feasible trajectory geometry. Let $\mathcal{M}$ denote the set of locally feasible target-following trajectories. If the expert trajectory satisfies $x_1\in\mathcal{M}$, then x-pred gives
\begin{equation}
\mathrm{dist}(\hat{x}_1,\mathcal{M})
\le
\|\hat{x}_1-x_1\|.
\end{equation}
Therefore, x-pred error directly controls distance to the feasible trajectory set. For an auxiliary parameterization, this distance is mediated by $\Psi_\lambda$, making the relation between prediction error and feasible-trajectory deviation less direct.

\subsection{Risk Reduction by Embedded Executable-Prefix Repair}

Let $\mathcal{C}_m(c)$ be the locally collision-free prefix set with margin $m$, and define the prefix violation functional
\begin{equation}
\rho(\tau,c)=\frac{1}{K_{\mathrm{exec}}}\sum_{k=1}^{K_{\mathrm{exec}}}\mathbf{1}\{\tau_k\notin\mathcal{C}_m(c)\}.
\end{equation}
For a trajectory point $q$ and local obstacle set $O(c)$, let $d(q,O)=\min_{o\in O(c)}\|q-o\|_2$ and define the hinge risk $r(q,c)=[m-d(q,O)]_+$. If RSEPSS moves a selected risky point $q$ along the outward unit vector from its nearest obstacle,
\begin{equation}
q'=q+\eta \frac{q-o^{\mathrm{nn}}}{\|q-o^{\mathrm{nn}}\|_2},
\qquad \eta\ge 0,
\end{equation}
and the nearest obstacle remains unchanged locally, then $d(q',O)\ge d(q,O)$ and hence $r(q',c)\le r(q,c)$. Thus each successful projection step does not increase local hinge risk.

Let $\eta_{\mathrm{rep}}$ denote the expected fraction of risky executable-prefix points that are triggered and successfully repaired, and let $\epsilon_{\mathrm{sm}}$ denote the residual fraction of violations caused by infeasible local geometry or smoothing. Then the repaired prefix satisfies
\begin{equation}
\mathbb{E}\bigl[\rho(S_m(\tau),c)\bigr]
\le
\mathbb{E}\bigl[\rho(\tau,c)\bigr]-\eta_{\mathrm{rep}}+\epsilon_{\mathrm{sm}}.
\label{eq:rsepss_repair_bound}
\end{equation}
Because our method applies $S_m$ inside the sampling update, the flow update uses the repaired clean trajectory rather than an unsafe candidate that will be corrected only after sampling. If the sampling update $U_s(x_s,\hat{x}_1)$ is Lipschitz in its clean endpoint with constant $L_U$, then replacing $\hat{x}_1$ by $S_m(\hat{x}_1)$ changes the next flow state by at most
\begin{equation}
\|U_s(x_s,S_m(\hat{x}_1))-U_s(x_s,\hat{x}_1)\|
\le
L_U\|S_m(\hat{x}_1)-\hat{x}_1\|.
\end{equation}
Thus embedded repair keeps the correction local and bounded while allowing the safer prefix to influence subsequent sampling states.

\subsection{OOD Risk Decomposition}

Let $P$ be the ID condition distribution and $Q$ an OOD condition distribution. Let $W_1(P_c,Q_c)$ denote the Wasserstein-1 distance between their condition marginals. Combining the preceding bounds yields the following decomposition for the OOD task risk:
\begin{align}
\mathcal{R}_Q(S_m\circ\Pi_\theta)
&=
\mathbb{E}_{c\sim Q}\left[\ell(S_m(\Pi_\theta(c,Z)),c)\right] \\
&\le
\mathcal{R}_P(S_m\circ\Pi_\theta)
+ L_c W_1(P_c,Q_c)
+ L_z\mathbb{E}_{Q}\|Z-Z^\star\| \\
&\quad
+ L_x\mathbb{E}_{Q}\|\Pi_\theta(c,Z^\star)-x_1^\star(c)\|
- \beta\eta_{\mathrm{rep}}
+ \epsilon_{\mathrm{sm}},
\label{eq:ood_risk_decomposition}
\end{align}
where $x_1^\star(c)$ is an ideal feasible tracking trajectory, $L_c$, $L_z$, and $L_x$ collect the relevant Lipschitz constants, and $\beta>0$ converts repaired prefix mass into task-risk reduction. The term $L_c W_1(P_c,Q_c)$ captures the OOD scene shift, the future-token term is controlled by the reliability-gated adapter in Equation~\ref{eq:future_adapter_bound}, the clean-trajectory error term is directly controlled by x-pred as in Equation~\ref{eq:xpred_task_risk}, and the last two terms capture the benefit and residual limitation of embedded RSEPSS from Equation~\ref{eq:rsepss_repair_bound}.

This analysis explains the qualitative pattern observed in the OOD tables. The method is not forced to minimize raw tracking error at all costs. Instead, it reduces the influence of unreliable target futures, keeps generation aligned with the clean feasible trajectory space, and repairs risky executable prefixes during sampling. As a result, under the hardest combined shift, the method may trade safe-tracking success against the MPC reference while retaining the low-collision and final-error behavior targeted by the learned safety-oriented planner.

\section{Safety and Parameterization Ablations}
\label{app:ablation}

This appendix reports the two module-level ablations that are most closely tied to the theoretical discussion in Appendix~\ref{app:robustness}: the embedded safety-refinement loop and the clean-trajectory flow parameterization. Both ablations are rerun on the same fixed benchmark as the main tables, with $M=256$ obstacle samples, 12 sampling steps, a fixed noise seed, and at most 25 replans per rollout. We report macro averages over the three ID obstacle-density groups and the four OOD shift groups.

\subsection{Embedded RSEPSS Ablation}
\label{app:rsepss_ablation}

Table~\ref{tab:app_rsepss_ablation} isolates the effect of embedding RSEPSS into the sampling loop. The no-RSEPSS rows correspond to the gated future-adapter planner before safety refinement, while the embedded-RSEPSS rows use the full method. RSEPSS removes nearly all dense collisions and raises safe tracking from 0.282 to 0.913 in ID and from 0.646 to 0.854 in OOD. The cost is a moderate increase in ADE/FDE and replanning time, which is expected because the refinement prioritizes executable safe prefixes over unconstrained geometric tracking.

\begin{table}[t]
\centering
\caption{Embedded RSEPSS ablation on the fixed ID/OOD benchmark. Metrics match the main tables and are macro-averaged over groups.}
\label{tab:app_rsepss_ablation}
\scriptsize
\setlength{\tabcolsep}{3.0pt}
\renewcommand{\arraystretch}{1.05}
\resizebox{\linewidth}{!}{
\begin{tabular}{llccccccc}
\toprule
\textbf{Split} & \textbf{Variant} & \textbf{ADE}$\downarrow$ & \textbf{FDE}$\downarrow$ & \textbf{CR}$\downarrow$ & \textbf{Smooth.}$\downarrow$ & \textbf{STT@8}$\uparrow$ & \textbf{STR@8}$\uparrow$ & \textbf{Time}$\downarrow$ \\
\midrule
ID
& gated future-adapter & \textbf{3.462} & \textbf{4.453} & 0.140 & 100.280 & 0.282 & 0.282 & \textbf{0.184} \\
& + embedded RSEPSS & 4.365 & 5.448 & \textbf{0.000} & \textbf{82.094} & \textbf{0.961} & \textbf{0.913} & 0.300 \\
\midrule
OOD
& gated future-adapter & \textbf{3.002} & \textbf{2.698} & 0.057 & 105.067 & 0.642 & 0.646 & \textbf{0.211} \\
& + embedded RSEPSS & 3.653 & 3.719 & \textbf{0.002} & \textbf{89.830} & \textbf{0.896} & \textbf{0.854} & 0.287 \\
\bottomrule
\end{tabular}}
\end{table}

\begin{figure}[t]
\centering
\begin{subfigure}[t]{0.48\linewidth}
    \caption{RSEPSS on}
    \includegraphics[width=\linewidth]{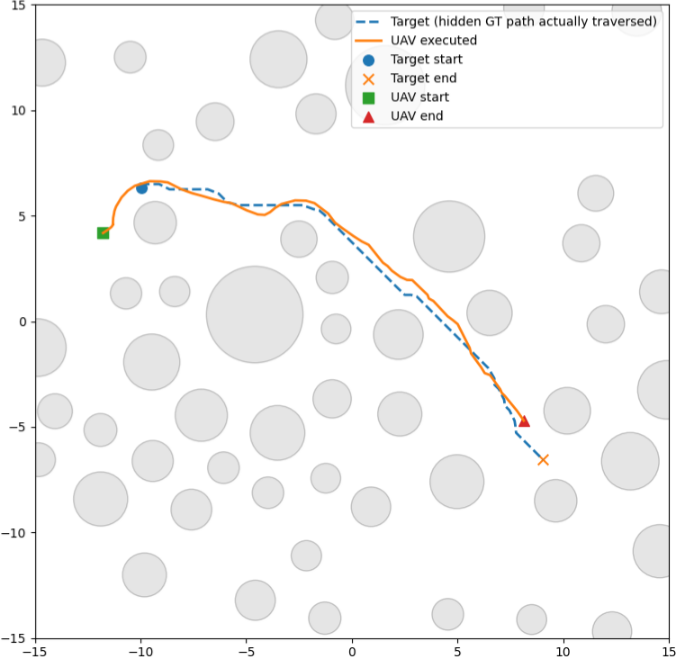}
\end{subfigure}
\hfill
\begin{subfigure}[t]{0.48\linewidth}
    \caption{RSEPSS off}
    \includegraphics[width=\linewidth]{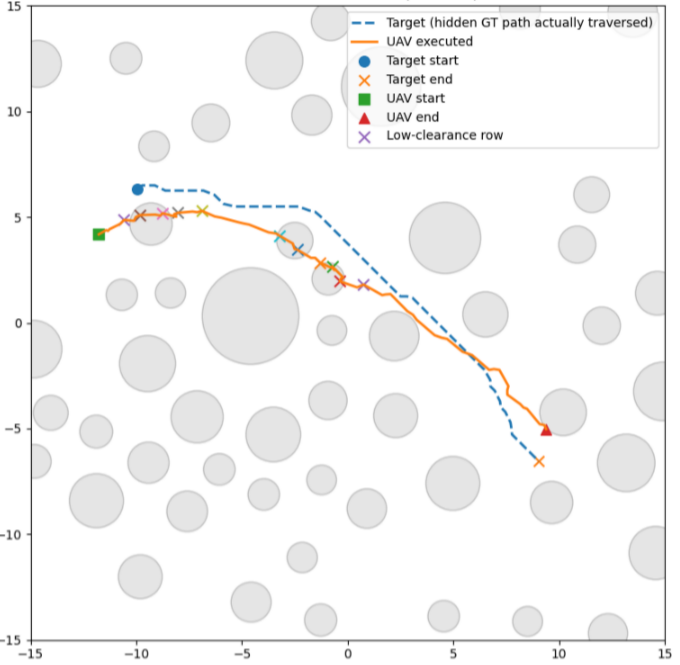}
\end{subfigure}
\caption{Qualitative effect of RSEPSS refinement. The embedded correction steers high-risk prefixes away from obstacles and yields executable trajectories with substantially higher safe-tracking rates.}
\label{fig:app_rsepss_vis}
\end{figure}

The qualitative rollouts in Figure~\ref{fig:app_rsepss_vis} show the same mechanism at the trajectory level. Without the embedded correction, the generated candidate can follow the target more directly but may enter a high-risk prefix near obstacles. Once RSEPSS is applied inside the sampling loop, the repaired prefix changes the subsequent denoising state rather than merely clipping the final output. This is the empirical counterpart of Equation~\ref{eq:rsepss_repair_bound}: the projection term reduces local obstacle risk by at least a margin-dependent amount, while the smoothing residual remains bounded. It also explains the OOD pattern in Equation~\ref{eq:ood_risk_decomposition}, where the safety-repair term reduces task risk even when the raw tracking term increases.

\subsection{x-pred versus v-pred Parameterization}
\label{app:xv_ablation}

Table~\ref{tab:app_xv_ablation} compares clean-trajectory x-pred against a velocity-style v-pred flow baseline. To keep the comparison focused, both variants use the same predictor conditioning and basic sampling interface, without RSEPSS or the gated future adapter. The x-pred parameterization substantially reduces tracking error and improves safe-tracking success, especially under OOD shifts. The v-pred baseline produces lower jerk, but this smoothness is coupled to poor tracking and low success; it therefore does not translate into the task-level safe-tracking objective.

\begin{table}[t]
\centering
\caption{x-pred versus v-pred parameterization ablation on the fixed ID/OOD benchmark. Both variants use the same predictor conditioning and no safety refinement.}
\label{tab:app_xv_ablation}
\scriptsize
\setlength{\tabcolsep}{3.0pt}
\renewcommand{\arraystretch}{1.05}
\resizebox{\linewidth}{!}{
\begin{tabular}{llccccccc}
\toprule
\textbf{Split} & \textbf{Variant} & \textbf{ADE}$\downarrow$ & \textbf{FDE}$\downarrow$ & \textbf{CR}$\downarrow$ & \textbf{Smooth.}$\downarrow$ & \textbf{STT@8}$\uparrow$ & \textbf{STR@8}$\uparrow$ & \textbf{Time}$\downarrow$ \\
\midrule
ID  & v-pred flow & 7.237 & 10.605 & \textbf{0.098} & \textbf{28.325} & \textbf{0.314} & 0.074 & 0.187 \\
ID  & x-pred flow & \textbf{3.947} & \textbf{5.553} & 0.119 & 103.361 & 0.307 & \textbf{0.309} & 0.187 \\
\midrule
OOD & v-pred flow & 7.891 & 10.292 & 0.069 & \textbf{24.961} & 0.350 & 0.146 & 0.192 \\
OOD & x-pred flow & \textbf{3.490} & \textbf{3.882} & \textbf{0.061} & 108.478 & \textbf{0.618} & \textbf{0.625} & \textbf{0.190} \\
\bottomrule
\end{tabular}}
\end{table}

This diagnostic supports the clean-trajectory design used by the final planner. In Equation~\ref{eq:xpred_task_risk}, the task loss depends directly on the error between the generated clean trajectory and the ideal feasible tracking trajectory. Predicting $x_1$ directly reduces this term, whereas a velocity parameterization must accumulate integration error across the sampling path. The ablation is deliberately narrower than the full method: it does not claim that x-pred alone solves safety. Instead, it shows why the base generator should operate in clean trajectory space, after which the gated future adapter controls future guidance and embedded RSEPSS controls the safety term in Equation~\ref{eq:ood_risk_decomposition}.

\end{document}